\documentclass[10pt,twocolumn,letterpaper]{article}
\usepackage[review]{cvpr}      
\newcommand{\hide}[1]{}
\usepackage[pagebackref]{hyperref}
\usepackage{graphicx}
\usepackage{amsmath}
\usepackage{amssymb}
\usepackage{amsfonts}
\usepackage{amsthm}
\usepackage{booktabs}
\usepackage{multicol}
\usepackage{multirow}
\usepackage{makecell}
\usepackage{comment}
\usepackage[utf8]{inputenc}
\usepackage{xcolor}
\usepackage{arydshln}
\usepackage{tikz}
\usepackage{pgfplots}
\usepackage{booktabs}
\usepackage{listings}
\usepackage[capitalize]{cleveref}
\crefname{section}{Sec.}{Secs.}
\Crefname{section}{Section}{Sections}
\Crefname{table}{Table}{Tables}
\crefname{table}{Tab.}{Tabs.}

\newcommand{\TP}[1]{\textcolor{red}{TP: #1}}
\newcommand{\tim}[1]{\textcolor{red}{TD: #1}}

\newcommand{\CC}{{\mathbb C}}

\newcommand{\QQ}{{\mathbb Q}}

\newtheorem{theorem}{Theorem}

\newtheorem{observation}[theorem]{Observation}
\theoremstyle{remark}

\theoremstyle{definition}
\newtheorem{definition}[theorem]{Definition}
\newtheorem{example}[theorem]{Example}

\usepackage[initials]{amsrefs} 
\def\confName{CVPR}
\def\confYear{2023}
\pgfplotsset{compat=1.18} 
\newif\ifArXiV
\ArXiVfalse
\begin{document}
\title{Camera Autocalibration Revisited}
\author{A P Dal Cin\\
\and
T Pajdla\\
}
\maketitle
\begin{abstract}
We present ...
\end{abstract}
\section{Introduction} \label{sec:intro}
Camera autocalibration ...

\section{Previous work}\label{sec:previous-work}

\begin{enumerate}

\item Triggs, B. (1998). Autocalibration from planar scenes. In: Burkhardt, H., Neumann, B. (eds) Computer Vision — ECCV'98. ECCV 1998. Lecture Notes in Computer Science, vol 1406. Springer, Berlin, Heidelberg. https://doi.org/10.1007/BFb0055661
\item Pollefeys, Marc, Koch, Reinhard, Van Gool, Luc. (1998). Self-Calibration and Metric Reconstruction in Spite of Varying and Unknown Internal Camera Parameters. Proceedings of the IEEE International Conference on Computer Vision. 90-95. 10.1109/ICCV.1998.710705. 
\item Gherardi Fuseillo, Practical Autocalibration 

\end{enumerate}




\section{Background}
Given $P$ cameras, all calibrated except for a common unknown focal length, and $N$ point correspondences. The degrees of freedom of the system are $6P + 3N - 7 + 1$, where the final term represents the unknown focal length of the camera, and the number of equations is $2NP$. 

\begin{figure}[h]
    \centering
    \includegraphics[width=\linewidth]{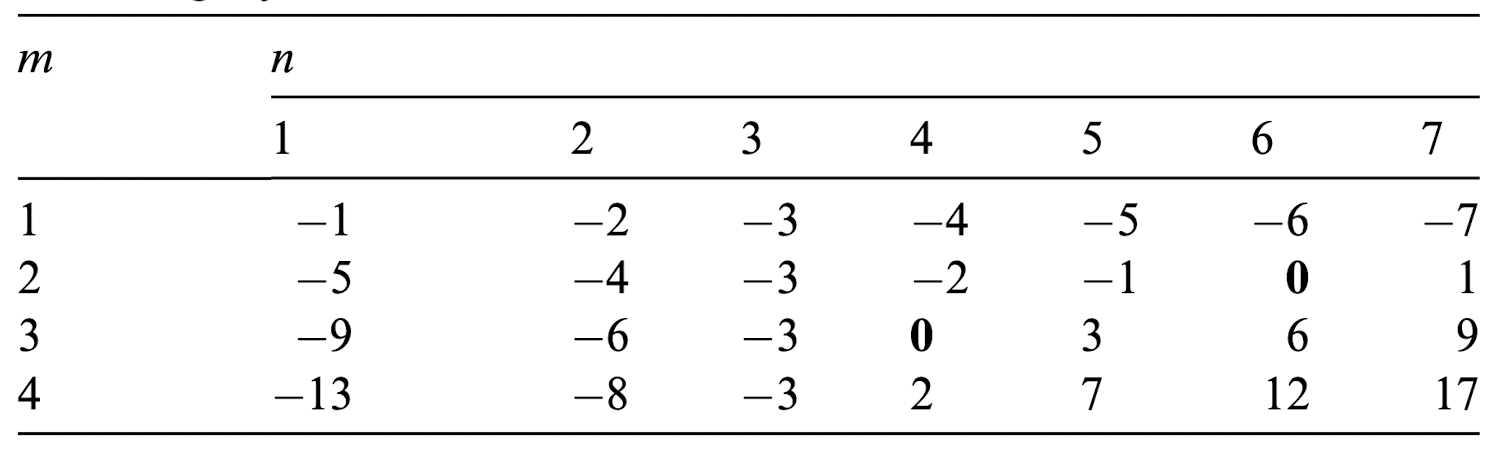}
    \caption{Number of excess constraints for $P$ cameras and $N$ points with unknown focal length $f$.}
    \label{fig:tab-p-vs-n-constraints}
\end{figure}

The minimal cases are thus $(P,N) = (2,6)$, for which several solutions have been proposed \cite{stewenius2008minimal}, and $(P,N) = (3,4)$, which is still an open problem.

\section{Two-view 6pt}
Consider $P = 2$ cameras, all calibrated except for a common focal length $f$, and $N=6$ image point correspondences. The f+E+f problem consists in estimating the relative pose $(\mathbf{R},\mathbf{t})$ and common unknown focal length $f$ of the two cameras. It was first solved by \cite{stewenius2008minimal} using a Gröbner basis solver. 

\subsection{Geometric constraints}
The fundamental matrix $\mathbf{F} = [f_{ij}]^3_{i,j=1}$ and corresponding image points $x$ and $x'$ satisfy the constraint:
\begin{equation}
    \mathbf{x}^T \mathbf{F} \mathbf{x}' = 0.
\end{equation}
$\mathbf{F}$ is a rank-2 matrix.

The fundamental matrix $\mathbf{F}$ satisfies:
\begin{equation}
    \label{eq:fund-essential}
    \mathbf{E} = \mathbf{K}^T \mathbf{F} \mathbf{K} = \mathbf{K} \mathbf{F} \mathbf{K}
\end{equation}
where $\mathbf{K} = \text{diag}(f,f,1)$ is the calibration matrix with unknown focal length $f$ and $\mathbf{E}$ is the essential matrix. The essential matrix $\mathbf{E}$ has the additional property that two non-zero singular values are equal \tim{Over $\mathbb{R}$}. This constraint is used in the context of self-calibration in \cite{mendoncca1999simple} to derive the non-linear Mendon\c{c}a-Cipolla cost function, which is commonly used in a Levenberg-Marquardt optimization scheme for the estimation of $\mathbf{K}$ from a known set of fundamental matrices $\{ \mathbf{F}_i \}$.

The essential matrix $\mathbf{E}$ has rank 2 and satisfies the Demazure equations (trace constraint):
\begin{equation}
    \label{eq:trace-constraint}
    2 \mathbf{E} \mathbf{E}^T \mathbf{E} - \text{trace}(\mathbf{E} \mathbf{E}^T)\mathbf{E} = 0.
\end{equation}

The linear equations from the epipolar constraints:
\begin{equation}
    \mathbf{x}_i^T \mathbf{F} {\mathbf{x}'}_i^T = 0
\end{equation}
are rewritten to:
\begin{equation}
    \mathbf{M} \mathbf{f} = 0,
\end{equation}
where $\mathbf{M}$ is a $6 \times 9$ coefficient matrix and $\mathbf{f}$ is a vector of 9 elements of the fundamental matrix $\mathbf{F}$. For six correspondences in two views, the coefficient matrix $\mathbf{M}$ has a three-dimensional null space. $\mathbf{F}$ can be parameterized as:
\begin{equation}
    \mathbf{F} = x \mathbf{F}_1 + y \mathbf{F}_2 + \mathbf{F}_3.
\end{equation}
By combining Eq.~\ref{eq:trace-constraint} and Eq.~\ref{eq:fund-essential}, we obtain ten third- and fifth-order polynomial equations in the unknowns $x, y$ and $w = 1/f^2$:
\begin{equation}
    2 \mathbf{F} \mathbf{Q} \mathbf{F}^T \mathbf{Q} \mathbf{F} - \text{trace}(\mathbf{F} \mathbf{Q} \mathbf{F}^T \mathbf{Q}) \mathbf{F} = 0.
\end{equation}
These equations are commonly solved in Gröbner basis solvers to the f+E+f problem \cite{}.

\begin{table*}[t]
\centering
\resizebox{\textwidth}{!}{
\begin{tabular}{@{}llllllll@{}}
\toprule
Paper              & $N$ & $\mathbf{R}$ & $\mathbf{t}$ & $f$ & $u$ & $v$ & Comments \\ \midrule
Some results on minimal euclidean reconstruction from four points \cite{quan2006some} & 3 & Yes & Yes & No & No & No & Gröbner basis solver \\
Fabbri, et al. \cite{fabbri2019trifocal} & 3 & Yes & Yes & No  & No & No & \emph{i}) 3 points and 1 line, \emph{ii}) 3 points, 2 lines through two of the points \\
Revisiting Trifocal Tensor Estimation using Lines \cite{kuang2014revisiting} & 3 & Yes & Yes & Yes & No & No & \emph{i}) 10 lines, \emph{ii}) 11 lines in 3 views\\
Minimal problems for the calibrated trifocal variety \cite{kileel-trifocal} & 3 & Yes & Yes & No & No & No & PPP, PPL, PLP, LLL, PLL \\ \bottomrule
\end{tabular}
}
\caption{3-view minimal problems for relative pose, focal length, principal point estimation. \TP{Zuzana had a paper where the principal point was involved: \url{https://openaccess.thecvf.com/content_cvpr_2018/papers/Larsson_Camera_Pose_Estimation_CVPR_2018_paper.pdf}}}
\vspace{1em}
\end{table*}

\section{Three-view 4pt (Scranton)}
\label{sec:scranton}
For Euclidean reconstruction, a generic system of $N$ points visible in $P$ calibrated images yields $2NP$ constraints in $3P + 6N - 7$ unknowns, where $7$ are the d.o.f. in the 3D coordinate system (including the scale of of the scene). Given that each 2D point provides $2$ constraints on the reconstruction, the following relation holds:
\begin{equation}
    \label{eq:2npge}
    2NP \ge 3P + 6N - 7.
\end{equation}

For 3-view Euclidean reconstruction ($N = 3$), a minimum of $P=4$ is required from unknown 3D points. From Eq.~\ref{eq:2npge}, $2NP = 24 \ge 3P + 6N - 7 = 23$. The polynomial system is overconstrained and has no solutions when the input data contains noise. In \cite{quan2006some}, \TP{The first solution to the overconstrained 4 points in 3 views was by Nister and Schaffalitzky from ECCV 2004, later published in IJCV: https://www.robots.ox.ac.uk/~vgg/publications/papers/nister05.pdf. Before, there were several other appears formulating and partially solving it. Many references are in Petr Hruby CVPR 2022.} the reconstruction problem from $4$ points in $3$ calibrated views is formulated based on Euclidean point-point and point-camera distances. The proposed formulation assumes a known camera calibration matrix $\mathbf{K}$ common across the $3$ views and normalized homogeneous image coordinates $\mathbf{u} = (u, v, 1)^T$. The 3D point $\mathbf{x}$ corresponding to $\mathbf{u}$ is determined by: \emph{i}) the 3D direction $\mathbf{K}^{-1}\mathbf{u}$ and \emph{ii}) 3D depth $\lambda$ as $\lambda \mathbf{x}$.

\paragraph{Euclidean Constraint}
The distance between two 3D points $\mathbf{p}$ and $\mathbf{q}$ is given by:
\begin{equation}
    || \mathbf{p} - \mathbf{q} ||^{2} = ||\mathbf{p}||^2 + ||\mathbf{q}||^2 - 2\, \mathbf{p}^T \mathbf{q}.
\end{equation}

Given the normalized direction vectors representing the 3D points in the camera frame and given $||\mathbf{x}_p|| = 1$:
\begin{equation}
    \delta_{pq}^2 = \lambda_p^2 + \lambda_q^2 - c_{pq} \lambda_p \lambda_q
\end{equation}
where $c_{pq} = 2\mathbf{x}_i^T \mathbf{x}_{j} = 2 \text{cos}(\theta_{pq})$ is known from the image points. $\delta_{pq}$ is the unknown distance between the space points.

\paragraph{Polynomial system}
For $N=3$ images of $P=4$ points, we obtain a system of $18$ homogeneous polynomials in $12$ unknowns $\lambda_{ip}$ and $6$ unknowns $\delta_{ij}$:
\begin{equation}
    f(\lambda_{ip},\lambda_{iq},\delta_{pq})=0
\end{equation}
where $i = 1, \ldots, N$ and $p < q = 1, \ldots, 4$. However, the unknown inter-point distances $\delta_{ij}$ can be eliminated as follows:
\begin{equation}
    \delta_{pq}^2 = \lambda_{ip}^2 + \lambda_{iq}^2 - c^{i}_{pq} \lambda_{ip} \lambda_{iq} = \lambda_{jp}^2 + \lambda_{jq}^2 - c^{j}_{pq} \lambda_{jp} \lambda_{jq}
\end{equation}
leaving $12$ homogeneous polynomials in $12$ unknowns $\lambda_{ip}$. We can also de-homogenize to obtain 11 inhomogeneous unknowns.

\paragraph{Symmetric dehomogenization} The unknowns are depths $\lambda$ of 3D point relative to the camera center. If $\lambda_{ip}$ is a solution, then $-\lambda_{ip}$ is also a solution. The polynomial system can be de-homogenized by introducing relative depths between points of the same image:
\begin{equation}
    y_{ip} = \frac{\lambda_{ip}}{\lambda_{i1}}, \; \; i \in \{1, 2, 3\}, p = 1, \ldots, 4.
\end{equation}
In addition, we obtain relative scales for images:
\begin{equation}
    z_j = \frac{\lambda_{j1}^2}{\lambda_{11}^2} , \; \; j \in \{2, 3\}.
\end{equation}

The de-homogenized ideal $<g(y_{ip},z_j)>$ is obtained by fixing scales $y_{i1} = 1$ and $z_{1} = 1$. $g$ has $11$ inhomogeneous unknowns. This transformation eliminates the quadratic symmetry solutions:
\begin{equation}
    \lambda_{ip} \hookrightarrow -\lambda_{ip}
\end{equation}
This process is called \emph{symmetric dehomogenization} as the same point is used in different images for normalization. In \cite{quan2006some}, the authors explain that this formulation contains many spurious solutions.

\paragraph{Asymmetric dehomogenization}
The relative depths for points in the same image become:
\begin{equation}
    y_{ip} = \frac{\lambda_{ip}}{\lambda_{ii}}, \; \; i \in \{1, 2, 3\}, p = 1, \ldots, 4.
\end{equation}
and relative scales (squared)
\begin{equation}
    z_j = \frac{\lambda_{j1}^2}{\lambda_{11}^2}, \; \; j = \{2, 3\}
\end{equation}
We obtain a new dehomogenized ideal $<g'(y_{ip},z_j)>$, also with $11$ inhomogeneous unknowns.

\paragraph{Relaxation} We obtain a minimal problem by replacing $\mathbf{u}_{1,1}$ by $\mathbf{u}_{1,1} + \mathbf{l}[0;1;0]$, where $\mathbf{l}$ is a new unknown.  \tim{Should $\mathbf{l}$ be bold, since it is a scalar?}
This relaxation allows to adjust the second coordinate of the first 2D point in the first camera. Thus, the equations involving the first point in the first view, which has $\lambda_{1,1} = 1$, become:
\begin{equation}
    ||\mathbf{u}_{1,1} + l[0;1;0] - \lambda_{1,q} \mathbf{u}_{1,q}||^2 = ||\lambda_{1,2} \mathbf{u}_{1,2} - \lambda_{2,q} \mathbf{u}_{2,q}||^2
\end{equation}
Thus, we obtain $12$ de-homogenized polynomials in $11$ unknown depths $\lambda$ and the unknown $\mathbf{l}$.

\paragraph{Coplanarity} Four points in space may be coplanar. In \cite{quan2006some}, the authors argue that the Euclidean constraint still holds for coplanar points.

\subsection{Contribution}

\section{Previous work}\label{sec:sota}

\section{3V4P with unknown focal length}
Consider the matrix $\mathbf{K}$ of internal camera parameters:
\begin{equation}
    \mathbf{K} = \begin{bmatrix} f & 0 & 0 \\ 0 & f & 0 \\ 0 & 0 & 1 \end{bmatrix}
\end{equation}
and its inverse $\mathbf{K}^{-1}$ with $\tilde{f} = 1/f$.
\begin{equation}
    \mathbf{K}^{-1} = \begin{bmatrix} 1/f & 0 & 0 \\ 0 & 1/f & 0 \\ 0 & 0 & 1 \end{bmatrix} = \begin{bmatrix} \tilde{f} & 0 & 0 \\ 0 & \tilde{f} & 0 \\ 0 & 0 & 1 \end{bmatrix}
\end{equation}
The 3D direction of the 2D point $\mathbf{u} = [u, v, 1]^{T}$ is $\mathbf{K}^{-1} \mathbf{u}$ and the corresponding 3D point $\mathbf{x}$ is given by:
\begin{equation}
\label{eq:reprojection}
    \mathbf{x} = \lambda \mathbf{K}^{-1} \mathbf{u}
\end{equation}

Consider the Euclidean constraint:
\begin{equation}
    \delta_{pq}^2 = \lambda_{ip}^2 + \lambda_{iq}^2 - c^{i}_{pq} \lambda_{ip} \lambda_{iq} = \lambda_{jp}^2 + \lambda_{jq}^2 - c^{j}_{pq} \lambda_{jp} \lambda_{jq}.
\end{equation}

For $N=3$ images, we obtain 18 homogeneous polynomials $f(\lambda_{ip}, \lambda_{iq}, \delta_{pq}, \tilde{f})$ in 19 unknowns $\lambda_{ip}, \delta_{ij}, \tilde{f}$, or, equivalently, 12 homogeneous equations in 13 unknowns $\lambda_{ip}, \tilde{f}$. We remove the 3D scale factor (de-homogenize) by setting $\lambda_{1,1} = 1$ and obtain $12$ inhomogeneous equations in $12$ unknowns $\lambda_{ip}, \tilde{f}$, with $(i,p) \neq (1,1)$.

We obtain 1326 complex solutions and 68 real solutions.

\section{Uncalibrated zero-skew 3V5P}
We formulate the uncalibrated zero-skew 3V5P problem by considering the matrix $\mathbf{K}$ of intrinsic camera parameters with zero-skew:
\begin{equation}
    \mathbf{K} = \begin{bmatrix}
        f_x & 0 & c_x \\
        0 & f_y & c_y \\
        0 & 0 & 1
    \end{bmatrix}
\end{equation}
where $f_x, f_y$ are the focal lengths of the camera and $(c_x, c_y)$ are the coordinates of the principal point. In the aforementioned uncalibrated case, the system of 5 points in 3 views has $3P + 6N - 7 + 4 = 30$ unknowns, where $4$ are the d.o.f. introduced by the unknown calibration parameters. In the projective depth formulation introduced in \cite{quan2006some}, the number of unknowns is reduced to $15$ unknown projective depths $\lambda_{ip}$ and $4$ calibration parameters in $\mathbf{K}$, for a total of $19$ unknowns. After de-homogenization, i.e., setting $\lambda_{11} = 1$ and scaling the other projective depths $\lambda_{ip}$ accordingly, we obtain $18$ unknowns.

Following the rationale in the original Scranton problem (Sec~\ref{sec:scranton}), we derive $2 \times {{5}\choose{2}} = 20$ equations. As discussed in \cite{quan2006some}, the structure parametrization with inter-point distances $\delta_{pq}$ is effective in the 3V4P case, as the ${4}\choose{2}$  distances correspond to the edge length of the tetrahedron in Fig.~2 of \cite{quan2006some}. For $N \geq 5$ points, not all the ${N}\choose{2}$ inter-point distances are required to enforce the rigidity of the system. Instead, we can consider any $4$ points from the set $\{ \mathbf{x}_i \}_{i=1}^N$ to build a tetrahedron, leading to $2 \times 6 = 12$ independent constraints. The remaining 3D points can be rigidly attached to the tetrahedron by considering exactly $3$ inter-point distances. As a result, each 3D point gives $3 \times 2 = 6$. The generic formula for computing the number of independent equations given by $N \geq 4$ points observed in $3$ views is the following:
\begin{equation}
\label{eq:independent-eqs}
    \mathrm{\#eqs} = 2 \times \left[ {{4}\choose{2}} + (N - 4) \times 3 \right]
\end{equation}

In our experiments, we ignore the inter-point distance between $\mathbf{x}_4$ and $\mathbf{x}_5$ in order to obtain a system of $18$ independent equations. We apply symmetric de-homogenization (Sec.~\ref{sec:scranton}) and obtain 8779 complex solutions and 88 real solutions, 44 of which are valid (positive focal lengths and principal point coordinates).

\section{Uncalibrated 3V6P}
We formulate the uncalibrated 3V6P problem by considering the matrix $\mathbf{K}$ of intrinsic camera parameters:
\begin{equation}
    \mathbf{K} = \begin{bmatrix}
        f_x & s & c_x \\
        0 & f_y & c_y \\
        0 & 0 & 1
    \end{bmatrix}
\end{equation}

We obtain a system with $3P + 6N - 7 + 5 = 18 + 18 - 7 + 5 = 34$ unknowns, where $5$ are the d.o.f. introduced by the unknown calibration matrix. From 2D point coordinates we can derive a polynomial system of $2NP = 2 \times 3 \times 6 = 36$ equations, meaning that the problem is overconstrained.

If we apply the depth formulation of \cite{quan2006some}, the number of unknowns projective depths $\lambda_{ip}$ is $3 \times 6 = 18$. Considering the matrix $\mathbf{K}$ of intrinsic parameters, the total number of homogeneous unknowns is $23$, whereas, after dehomogenization, we have $22$ unknowns.

Following Eq.~\ref{eq:independent-eqs}, we can derive $24$ independent constraints in terms of the inter-point distances between the points $\{ x_i \}_{i=1}^{P}$. If we include in the computation even those constraints that are not independent, we obtain $2 \times {{6}\choose{2}} = 2 \times 15 = 30$ equations. It follows that ${{30}\choose{8}} = 5852925$ are the possible combinations of constraints removed from the system in order to have an equal number of equations and unknowns.

In our experiments, we follow a similar strategy to the 3V5P problem, and consider the tetrahedron described by the inter-point distances between the points $\{ x_i \}_{i \in \{ 1, \ldots 4\} }$. Then, for $\mathbf{x}_5$, we do not consider its distance w.r.t. $\mathbf{x}_3$. Instead, for $\mathbf{x}_6$, we do not consider its distance w.r.t. $\mathbf{x}_4$. Similarly to 3V5P, we do not consider the distance between $\mathbf{x}_5$ and $\mathbf{x}_6$. Finally, to reach the desired $22$ equations, we discard the constraint given by the distance between $\mathbf{x}_1$ and $\mathbf{x}_5$, only for view-pair $1-2$. Also, we discard the constraint given by the distance between $\mathbf{x}_2$ and $\mathbf{x}_6$, only for view-pair $2-3$.

We apply symmetric dehomogenization (Sec.~\ref{sec:scranton}) and obtain 38921 complex solutions and 122 real solutions.

\section{Combinatorics}
The number of unknowns in the matrix of intrinsic parameters $\mathbf{K}$ determines the number of equations in the polynomial system and, hence, the number of points seen by the $N$ cameras. Depending on the number of unknown parameters, some constraints derived from the inter-point distances between the 3D points need to be omitted from the polynomial system. 

\begin{table}[tb]
\centering
\begin{tabular}{@{}lccccc@{}}
\toprule
\# K & V & P & \# Edges & \# Constr & \# Params \\ \midrule
1   & 3 & 4 & 6   & 12        & 12        \\
2   & 3 & 5 & 10  & 20        & 16        \\
3   & 3 & 5 & 10  & 20        & 17        \\
4   & 3 & 5 & 10  & 20        & 18        \\
5   & 3 & 6 & 15  & 30        & 22        \\ \bottomrule
\end{tabular}
\caption{Abc}
\label{tab:combinatorics-overview}
\end{table}

In addition, the projective depth parameterization of the problem is convenient for the 3V4P, but not as much for the 3V5P and 3V6P, as the inter-point distances are not all independent. As a result, some combinations of constraints result in unfeasible Euclidean reconstruction and calibration. In this section, we address the problem of identifying which constraint combinations lead to an underconstrained polynomial system for a given arbitrary number of unknown intrinsic parameters.

\subsection{Equivalence Classes for Point-Graphs}

\subsection{3V5P}

\medskip
\noindent
\textbf{4 unknowns in $\mathbf{K}$:} In this configuration, 2 constraints need to be removed from the 20 available.  

\newpage
\onecolumn
\newcommand{\draft}[1]{}
\newcommand{\R}{\mathbb R}
\newcommand{\C}{\mathbb C}
\newcommand{\arr}[2]{\begin{array}{#1} #2\end{array}}
\newcommand{\mat}[2]{\left[\!\!\arr{#1}{#2}\!\!\right]}
\renewcommand{\vec}{{}}
\newcommand{\M}{{}}
\section{Camera auto-calibration}
Camera auto-calibration can be formulated as solving projection equations for camera calibration matrices $\M{K}_i$. In the next, $i$ will index images and $j$ will index points. Image projections $\vec{x}_{ij\beta_i}$ are measured in the respective bases $\beta_i$ of images $i$ and the coordinates of vectors $\vec{X}_{j\delta}$ and camera centers $\vec{C}_{i\delta}$ are in a world coordinate system $\delta$. We will drop subscripts $\beta_i$ and $\delta$ to simplify our notation. 

With this, the projection equations read as 
\begin{eqnarray}
\eta_{ij}\,\vec{x}_{ij} = 
\M{K}_i\,\M{R}_i\mat{cc}{\M{I} &-\vec{C}_i}\,\mat{c}{\vec{X}_j\\1}, && i=1,\ldots,M,\ j=1,\ldots,N \label{eq:zx=K[R|-RC][X;;]}\\
\M{K}_i=\mat{ccc}{k_{1i}&k_{3i}&k_{4i}\\0&k_{2i}&k_{5i}\\0&0&1}, &&
\eta_{ij} \neq 0,\ \det{\M{K}_i} \neq 0,\ \M{R}_i^\top\M{R}_i=\M{I},\ \det{\M{R}_i} = 1 \label{eq:K=[...oool]}
\end{eqnarray}
The above equations comprise $M\,N$ unknowns $\eta_{ij}$, $5\,M$ unknowns $\M{K}_i$, $9\,M$ unknowns $\M{R}_i$, $3\,M$ unknowns $\vec{C}_j$, and $3\,N$ unknowns $\vec{X}_j$. Altogether, the Zariski closure~\cite{Cox-IVA-2015} of the set of solutions satisfying Equations~\ref{eq:zx=K[R|-RC][X;;]},~\ref{eq:K=[...oool]} is a variety ${\mathcal V}$ in $\C^{M\,N+17\,M+3\,N}$ affine space. 

Let us investigate the set of solutions of~\ref{eq:zx=K[R|-RC][X;;]} and~\ref{eq:K=[...oool]}.
\paragraph{Homogeneity}
Equations~\ref{eq:zx=K[R|-RC][X;;]} are homogeneous in $\eta_{ij}$, $\vec{C}_i$, and $\vec{X}_j$. Thus, if $\eta_{ij}$, $\vec{C}_i$, and $\vec{X}_j$ is a solution, then $\alpha\,\eta_{ij}$, $\alpha\,\vec{C}_i$, and $\alpha\,\vec{X}_j$ is also a solution for every $\alpha \in \C \setminus \{0\}$. 
\draft{
The homogeneity pattern may be very complicated for non-generic situations. For instance, if $M=N=2$ and  $\vec{C}_i = \vec{X}_i$, then we have equations
\begin{eqnarray}
    \eta_{11} \vec{x}_{11} &=& \M{K}_{1}\M{R}_1\mat{cc}{\M{I}&-\alpha_1\vec{C}_1} \mat{c}{\alpha_1\vec{C}_1\\1} = 0 \implies \eta_{11}=0,\ \forall \alpha_1\\
    \beta_1\beta_2\eta_{12} \vec{x}_{12} &=& \M{K}_{1}\M{R}_1\mat{cc}{\M{I}&-\beta_1\vec{C}_1} \mat{c}{\beta_2\vec{C}_2\\1} \\
    \beta_2\beta_1\eta_{21} \vec{x}_{21} &=& \M{K}_{2}\M{R}_2\mat{cc}{\M{I}&-\beta_1\vec{C}_2} \mat{c}{\beta_2\vec{C}_1\\1}\\
    \eta_{22} \vec{x}_{22} &=& \M{K}_{2}\M{R}_2\mat{cc}{\M{I}&-\alpha_2\vec{C}_2} \mat{c}{\alpha_2\vec{C}_2\\1} = 0 \implies \eta_{22}=0,\ \forall \alpha_2
\end{eqnarray}
and thus a solution $\beta_1\beta_2\eta_{ij}, $}
\paragraph{Geometric classes of solutions}
We see that 
    \begin{eqnarray*}
        \underbrace{\left(\frac{\eta_{ij}}{ \tau_i\, \psi_j}\right)}_{\eta_{ij}'} \,\vec{x}_{ij}  &=& 
        \underbrace{\left(\frac{\M{K}_i\,\M{R}_i\,\mat{cc}{\M{I} &-\vec{C}_i}\,\M{H}^{-1}}{\tau_i
        }\right)}_{(\M{K}_{i\M{s}}'\,\M{S})\,(\M{S}\,\M{R}_{i\M{s}}')\,\mat{cc}{\M{I} &-\vec{C}_i'}} 
        \underbrace{\left(\frac{\M{H}}{\psi_i}\,\mat{c}{\vec{X}_j\\1}\right)}_{\mat{c}{\vec{X}_j'\\1}} \\
        \mbox{with}\quad \tau_i &=& \left\| \mat{cc}{\M{r}_{i}^\top &-\M{r}_{i}^\top\vec{C}_i}\,\M{H}^{-1}(1:4,1:3) \right\| \neq 0,\ \mbox{where}\ \M{r}_{i}^\top\ \mbox{is the third row of}\ \M{R}_i\\
        \psi_j &=& \M{a}^\top\vec{X}_j+c \neq 0 \\
        \mbox{and}\quad \M{S} &=& \diag{\mat{ccc}{s&s&1}},\ \mbox{where}\ s \in \{-1,1\}
    \end{eqnarray*}
    for a regular $\M{H} = \mat{cc}{\M{A}&\M{b}\\\M{a}^\top&c} \in \C^{4 \times 4}$ with a regular $\M{A} \in \C^{3 \times 3}$. Quantities $\tau_i$ ensure $\|\M{P}_i'(3,1:3)\|=1$ for $\M{P}_i'=\left(\M{K}_i\,\M{R}_i\,\mat{cc}{\M{I} &-\vec{C}_i}\,\M{H}^{-1}\right)/\tau_i$, and thus $\M{P}_i'$ can be decomposed into a product of a K-matrix with $\M{K}(3,3)=1$ and a rotation. For every $\M{H}$, there are two decompositions of $\M{P}_i'$, one with $s=1$ and the other with $s=-1$. Quantities $\psi_j$ ensure that the 3D points remain in the chosen affine chart.
    
    Thus, for every solution $\sigma = (\eta_{ij},\M{K}_i, \M{R}_i,\vec{C}_i,\vec{X}_j)$ we have a set of solutions
    \begin{equation*}
       \Sigma(\sigma) = \left\{(\eta_{ij}',\M{K}_{is}', \M{R}_{is}',\vec{C}_i',\vec{X}_j') \mid   
        \M{H} \in \C^{4 \times 4},\ |\M{H}|\neq 0,\ |\M{A}|\neq 0,\ 
        1-\M{a}^\top\M{A}^{-1}\vec{C}_i \neq 0,\
        \tau_i \neq 0,\ \psi_j \neq 0,\ s \in \{-1,1\} 
        \right\}
    \end{equation*}
that are generated by geometrical transformations $\M{H}$\footnote{We have $|\M{A}| \neq 0$ and $1-\M{a}^\top\M{A}^{-1}\vec{C}_i \neq 0$ from the following. $\M{K}_i'\M{R}_i'\mat{cc}{\M{I}&-\vec{C}_i'} = \M{K}_i\M{R}_i\mat{cc}{\M{I}&-\vec{C}_i}\mat{cc}{\M{A}&\M{b}\\\M{a}^\top&c}^{-1}$ implies 
$\M{K}_i'\M{R}_i'\mat{cc}{\M{I}&-\vec{C}_i'}\mat{cc}{\M{A}&\M{b}\\\M{a}^\top&c} = \M{K}_i\M{R}_i\mat{cc}{\M{I}&-\vec{C}_i}$, and thus $\M{K}_i'\M{R}_i'\left(\M{A}-\vec{C}_i\,\M{a}^\top\right) = \M{K}_i\M{R}_i$. Hence, $|\M{K}_i'|\,|\M{R}_i'|\,|\left(\M{A}-\vec{C}_i\,\M{a}^\top\right)| = |\M{K}_i|\,|\M{R}_i|$ and thus $0 \neq |\M{K}_i'||\M{R}_i'| =1/|\left(\M{A}-\vec{C}_i\,\M{a}^\top\right)| = |\M{K}_i|/\left(|\M{A}|\, \left(1-\M{a}^\top\M{A}^{-1}\vec{C}_i \right) \right)$.}.
\paragraph{Equivalent solutions}
If $\M{A} \in SO(3)$ and $\M{a} = 0$, then choosing $\M{H}$ corresponds to choosing a scaled Cartesian coordinate system for $\vec{X}_j$. Similarly, choosing $\alpha$ chooses a particular scale for $\eta_{ij}, \vec{X}_j$. In these cases, the set of solutions in $\M{K}_i$ is invariant to the choices. In applications, we are interested in geometrical quantities, e.g., matrices $\M{K}_i$ or angles, that are invariant to the choice of the coordinate system and scaling. Hence, we consider two solutions $\sigma = (\eta_{ij},\M{K}_i, \M{R}_i,\vec{C}_i,\vec{X}_j)$, $\sigma' = (\eta_{ij}',\M{K}_i', \M{R}_i',\vec{C}_i',\vec{X}_j')$ equivalent , i.e., $\sigma \equiv \sigma'$,
if there are $c \in \C \setminus \{0\}$, $\M{A} \in SO(3)$, and $\M{b} \in \C^3$ such that 
\begin{equation}
     \eta_{ij}' = \eta_{ij} / c,\
     \M{K}_i' = \M{K}_i,\
     \M{R}_i' = \M{R}_i\,\M{A}^{-1},\     
     \vec{C}_i' = \left(\M{A}^{-1}\M{b}+\vec{C}_i\right)/c,\ 
     \vec{X}_j' = \left( \M{A}\,\vec{X}_j + b \right) / c
\end{equation}

Let us now look at three very classical auto-calibration problems. 
\begin{example}[Relative pose of two calibrated cameras]
Consider $M = 2$ calibrated cameras, i.e., $\M{K}_1 = \M{K}_2 = \M{I}$, and $N = 5$ points. The dimension of ${\mathcal V}$ is $d = 11\times 2 + 3\times 5 - 2\times 2\times 5 - 1 - K = 16 - K$. To reduce $d$ to zero, we need to add $K = 16 = 2\times 5 + 6$ constraints. Fixing $\M{K}_1 = \M{K}_2 = \M{I}$ gives $2\times 5$ linear constraints and setting $\M{R}_1 = \M{I}$, $\vec{C}_1=0$ adds $12$ linear constraints but also removes 6 constraints from $\M{R}_1^\top\M{R}_1 = \M{I}$, $\det{\M{R}_1} =1$, which are satisfied automatically.
\end{example}

\begin{example}[Relative pose of three cameras with the same unknown focal length]
    Consider $M = 3$ cameras, all calibrated up to a common unknown focal length, i.e., $\M{K}_1 = \M{K}_2 = \M{K}_2 = \diag{\mat{ccc}{f&f&1}}$, and $N = 4$ points. The dimension of ${\mathcal V}$ is $d = 11\times 3 + 3\times 4 - 2\times 3\times 4 - 1 - K = 20 - K$. To reduce $d$ to zero, we need to add $K = 20 = 3\times 5 - 1 + 6$ constraints. Fixing $\M{K}_1 = \M{K}_2 = \M{K}_2 = \diag{\mat{ccc}{f&f&1}}$ gives $3\times 5 - 1$ linear constraints and setting $\M{R}_1 = \M{I}$, $\vec{C}_1=0$ adds $12$ linear constraints but also removes 6 constraints from $\M{R}_1^\top\M{R}_1 = \M{I}$, $\det{\M{R}_1} =1$, which are satisfied automatically.
\end{example}

\begin{example}[Relative pose of two uncalibrated cameras]
Consider $M=2$ uncalibrated cameras, i.e., with unknown $\M{K}_1$, $\M{K}_2$ calibration matrices, and $N=7$ points. The dimension of ${\mathcal V}$ is $d = 11\times 2 + 3\times 7 - 2\times 2\times 7 - 1 - K = 14 - K$. To reduce $d$ to zero, we need to add $K = 14 = 12 + 6$ constraints. We notice that in this setup, with no constraints on the calibration matrices, $\M{H}^{-1}$ can change one of the projection matrices, e.g., $\M{K}_1\,\M{R}_1\,\mat{cc}{\M{I}&-\vec{C}_1}$, into any $\M{K}_1'\,\M{R}_1'\,\mat{cc}{\M{I}&-\vec{C}_1'}$ with $\M{K}_1'(3,3)=1$. Thus, it is possible to recover the calibration matrices and the scene only up to the action of $\M{H} \in GLN(4)$. We notice that to fix a representative in the set of solutions related by $GLN(4)$, we need to add 16 constraints, two more than previously counted. The difference is caused by two constraints on $\M{H}^{-1}$ following from that the squared norms of their third rows of $\M{K}_1'\M{R}_1'$ and $\M{K}_2'\M{R}_2'$ must equal to 1. Thus, we fix a single orbit representative by adding {\em independent} 14 constraints. The standard way~\cite{HZ-2003} of adding constraints is to set $\M{K}_1 = \M{R}_1 = \M{I}$ and $\vec{C}_1=0$. It adds only $11$ independent linear constraints since we automatically have $\M{I}(3,3)=1$. Thus, we need to add another three independent constraints next to the constraint on the norm of the third row of $\M{K}_2'\M{R}_2'$. 
\end{example}

The following example shows why we required $\eta_{ij} \neq 0$. 
\begin{example}[Spurious solutions]
    Allowing $\eta_{ij} = 0$ may lead to infinitely many spurious solutions. For instance, if $\eta_{11}=1$ but all other $\eta_{ij}=0$, then the following are solutions: $\M{K}_i=\M{I}$, $\M{R}_1=\M{I}$, $\vec{X}_1 = \vec{x}_{11}$, $\vec{X}_j = 0$ for all $j \neq 1$, $\vec{C}_i = 0$ and there are infinitely many $\M{R}_i$ for all $i \neq 1$ that leave the projection of $\vec{X}_1$ fixed in the images.
\end{example}

Let us next count the number of constraints that generically determine the dimension of ${\mathcal V}$. Equations~\ref{eq:zx=K[R|-RC][X;1]} reduce the dimension by $3\,M\,N$, constraints $\M{R}_i^\top\M{R}_i=\M{I}$, $\det{\M{R}_i}=1$ reduce the dimension by $6\,M$. Thus, altogether, we get the dimension $M\,N+17\,M+3\,N - 3\,M\,N - 6\,M = 11\,M + 3\,N - 2\,M\,N$. Hence, to reduce the dimension to $0$, i.e., to balance the problems, we need to add additional independent constraints $f_k(Z)=0$, $k=1,\ldots, K$ on unknowns $Z = (\eta_{ij},\M{K}_i, \M{R}_i,\vec{C}_i,\vec{X}_j)$. The formula for the dimension of ${\mathcal V}$ thus becomes
\begin{eqnarray}
    d({\mathcal V}) = 11\times M + 3\times N - 2\times M\times N - K - 1 
\end{eqnarray}

We see three issues when constructing constraints to reduce the dimension of ${\mathcal V}$ to zero: 
\begin{enumerate}
    \item We dehomogenize equations with a non-homogeneous equation $h(\eta_{ij},\vec{C}_i,\vec{X}_j) = 0$, e.g., $\eta_{11}=1$.
    \label{choice:1}
    \item We remove infinite orbits of solutions of subgroups of $GLN(4)$ by equations $g_k(\M{R}_i,\vec{C}_i) = 0$ and $f_l(\M{K}_i)=0$ that fix $\M{H}$, e.g., $\M{R}_1=\M{I},\vec{C}_1=0$. 
    Equations $f_l(\M{K}_i)=0$ also allow us to incorporate prior knowledge about the calibration matrices. \label{choice:2}
    \item We impose enough constraints $\eta_{ij}\neq 0$ to avoid spurious solutions due to $\eta_{ij}=0$. 
    \label{choice:3}
\end{enumerate}
The above choice~\ref{choice:2} replaces general constraints $f_k(Z)=0$, $k=1,\ldots, K$ by more specific constraints $g_k(\M{R}_i,\vec{C}_i) = 0,\ k=1\,\ldots,K$, and $f_l(\M{K}_i)=0,\ l=1,\ldots,L$. This is motivated by decoupling constraints on $\M{K}_i$ from other constraints in practical auto-calibration applications. 

The formula for the dimension of ${\mathcal V}$ now becomes
\begin{eqnarray}
    d({\mathcal V}) = 11\times M + 3\times N - 2\times M\times N - K - L - 1 
\end{eqnarray}
Note that the above formula must be applied with care since, first, all constraints must reduce the dimension of ${\mathcal V}$ and, secondly, we must not add any constraint that makes the ${\mathcal V}$ empty due to the inconsistency of equations. It also follows that $K \leq 6\,M$, and $L \leq 5\,M$, since adding additional independent equations would make the problem inconsistent. The following example illustrates typical pitfalls of using the formula in a particular case when we cannot auto-calibrate two cameras. 
\begin{example}[Impossibility to auto-calibrate unknown $\M{K}_1 = \M{K}_2$ from two generic views]
    Consider $M=2$ cameras with equal but unknown calibration matrix $\M{K}$ and $N$ points. Dehomogenize by $\eta_{11}=1$ and fix the Cartesian coordinate system by $\M{R}_1=\M{I}$ and $\vec{C}_1=0$. Remove spurious positive dimensional components by $\eta_{ij}\neq 0$. The projection equations now read as
    \begin{equation*}
    \eta_{1j}\,\vec{x}_{1j} =
    \M{K}\mat{cc}{\M{I} & 0}\,\mat{c}{\vec{X}_j\\1}, \quad
    \eta_{2j}\,\vec{x}_{2j} = 
    \M{K}\,\M{R}_2\mat{cc}{\M{I} &-\vec{C}_2}\,\mat{c}{\vec{X}_j\\1}, \quad
    \M{R}_2^\top\M{R}_2=\M{I},\ \det{\M{R}_2}=1 
    \end{equation*}
    We see that any regular $\M{H} = \mat{ll}{\M{L}&0\\\M{a}^\top&c}$ with upper triangular $\M{L}$ and $\M{L}(3,3)=1$ gives
    \begin{equation*}
    \eta_{1j}\,\vec{x}_{1j} =
    \M{K}\,\M{L}\mat{cc}{\M{I} & 0}\,\M{H}^{-1}
    \mat{c}{\vec{X}_j\\1},\
    \eta_{2j}\,\vec{x}_{2j} = 
    \M{K}\,\M{R}_2\mat{cc}{\M{L}-\vec{C}_2\M{a}^\top&-c\,\vec{C}_2}\,\M{H}^{-1}\mat{c}{\vec{X}_j\\1},\ 
    \M{H}^{-1} =\mat{cc}{\M{L}^{-1}&0\\\M{a}^\top\M{L}^{-1}c^{-1}&c^{-1}}
    \end{equation*}    
    Assuming $\eta_{11}=1$, we must have $\M{a}^\top\M{L}^{-1}c^{-1}\vec{X}_1+c^{-1}=1$ and we can introduce $\M{K}' = \M{K}\,\M{L}$ and $\vec{C}_2' = -c\,\M{K}\,\M{L}\,\vec{C}_2$. To get $\M{R}_2'$ we have equation $\M{K}'\M{R}_2' = \M{K}\,\M{R}_2(\M{L}-\vec{C}_2\M{a}^\top)$ giving $\M{R}_2' = \M{L}^{-1}\M{R}_2(\M{L}-\vec{C}_2\M{a}^\top)$ that yields $\M{I} = \left(\M{L}^{-1}\M{R}_2\,\M{L}-\M{L}^{-1}\M{R}_2\,\vec{C}_2\M{a}^\top\right)^\top\left(\M{L}^{-1}\M{R}_2\,\M{L}-\M{L}^{-1}\M{R}_2\,\vec{C}_2\M{a}^\top\right)$. The constraints $1 = \det{\M{R}_2'} = \det{\M{L}^{-1}}\det{\M{L}}(1-\M{a}^\top\M{L}^{-1}\vec{C}_2)$ implies $\M{a}^\top\M{L}^{-1}\vec{C}_2=0$. Thus, these constraints can cut the dimension of the $9$ dimensional orbit generated by $\M{H}$ by maximally $8=1+6+1$. Hence, for any number $N$ of points, the dimension of ${\mathcal V}$ will never drop below $1$.
\end{example}
Let us further consider the possibility of dropping some image measurements. That might lead to a complicated balancing scheme since removing image measurements may remove some unknowns too. To simplify the setting, we will eliminate all unknowns $\M{R}_i, \vec{C}_i, \vec{X}_j$ and remain with $\eta_{ij}, \M{K}_i$ only. We note that we may get a finite number of solutions after the elimination even when the original problem did not have a finite number of solutions.  

Our elimination removes equations $g_k(\M{R}_i,\vec{C}_i)=0$ and $\M{R}_i^\top\M{R}_i=\M{I}$, $\det{\M{R}_i}=1$ but leaves $h(\eta_{ij},\vec{C}_i,\vec{X}_j) = 0$ intact if it does not contain $\vec{C}_i$ and $\vec{X}_j$, i.e., if $h(\eta_{ij},\vec{C}_i,\vec{X}_j) =  h(\eta_{ij}) = 0$. Similarly, $f_l(\M{K}_i)=0$, $\det{\M{K}_i}\neq 0$, and $\eta_{ij} \neq 0$ will stay unchanged too. 

First, eliminating $\vec{C}_i$'s from Equations~\ref{eq:zx=K[R|-RC][X;1]} yields
\begin{align}
    \eta_{ij}\,\vec{x}_{ij} - \eta_{ik}\,\vec{x}_{ik} = \M{K}_i\,\M{R}_i\,\vec{X}_j - \M{K}_i\,\M{R}_i\,\vec{C}_i - \M{K}_i\,\M{R}_i\,\vec{X}_k + \M{K}_i\,\M{R}_i\,\vec{C}_i = \M{K}_i\,\M{R}_i\,(\vec{X}_j - \vec{X}_k), 
    \label{eq:ex-ex=kr(X-X)} \\ i=1,\ldots,M,\ j=1,\ldots,N,\ k=j+1,\ldots,N \nonumber
\end{align}
Next,  we eliminate $\M{R}_i$'s from Equations~\ref{eq:ex-ex=kr(X-X)} using $\M{R}_i^\top\M{R}_i=\M{I}$ as
\begin{eqnarray}
    \eta_{ij}\,\M{K}_i^{-1}\vec{x}_{ij} - \eta_{ik}\,\M{K}_i^{-1}\vec{x}_{ik} &=& \M{R}_i\,(\vec{X}_j - \vec{X}_k) \nonumber \\
    \|\eta_{ij}\,\M{K}_i^{-1}\vec{x}_{ij} - \eta_{ik}\,\M{K}_i^{-1}\vec{x}_{ik}\|^2 &=& \|\vec{X}_j - \vec{X}_k\|^2 \label{eq:||eKix-eKix||=||X-X||}
\end{eqnarray}
and then, using pairs of Equations~\ref{eq:||eKix-eKix||=||X-X||}, we eliminate $X_j$'s as
\begin{eqnarray*}
     \|\eta_{ij}\,\M{K}_i^{-1}\vec{x}_{ij} - \eta_{ik}\,\M{K}_i^{-1}\vec{x}_{ik}\|^2 &=& \|\vec{X}_j - \vec{X}_k\|^2 \\
    \|\eta_{mj}\,\M{K}_m^{-1}\vec{x}_{mj} - \eta_{mk}\,\M{K}_m^{-1}\vec{x}_{mk}\|^2 &=& \|\vec{X}_j - \vec{X}_k\|^2\\
    \|\eta_{ij}\,\M{K}_i^{-1}\vec{x}_{ij} - \eta_{ik}\,\M{K}_i^{-1}\vec{x}_{ik}\|^2 
    &=&
    \|\eta_{mj}\,\M{K}_m^{-1}\vec{x}_{mj} - \eta_{mk}\,\M{K}_m^{-1}\vec{x}_{mk}\|^2 \
\end{eqnarray*}
Similarly, using $\det R_i = 1$ and triplets $j_1,j_2,j_3$ of points, we get for $i=1,\ldots,M,\ j_1,j_2,j_3=1,\ldots,N,\ k=1,\ldots,N$
\begin{align}
    \det \left( \M{K}_i^{-1} \mat{ccc}{\eta_{ij_1}\vec{x}_{ij_1} - \eta_{ik}\vec{x}_{ik} &\eta_{ij_2}\vec{x}_{ij_2} - \eta_{ik}\vec{x}_{ik}&\eta_{ij_3}\vec{x}_{ij_3} - \eta_{ik}\vec{x}_{ik}} \right) = \det \left( \M{R}_i \mat{ccc}{\vec{X}_{j_1}-\vec{X}_k&\vec{X}_{j_2}-\vec{X}_k&\vec{X}_{j_3}-\vec{X}_k} \right) \nonumber \\
    \det(\M{K}_i^{-1})
    \det \mat{ccc}{\eta_{ij_1}\vec{x}_{ij_1} - \eta_{ik}\vec{x}_{ik} &\eta_{ij_2}\vec{x}_{ij_2} - \eta_{ik}\vec{x}_{ik}&\eta_{ij_3}\vec{x}_{ij_3} - \eta_{ik}\vec{x}_{ik}} = \det \mat{ccc}{\vec{X}_{j_1}-\vec{X}_k&\vec{X}_{j_2}-\vec{X}_k&\vec{X}_{j_3}-\vec{X}_k} \nonumber
\end{align}
and then by eliminating the right-hand sides we get for $i, m=1,\ldots,M,\ j_1,j_2,j_3=1,\ldots,N,\ k=1,\ldots,N$
\begin{multline*}
    \det(\M{K}_i^{-1})\,\det \mat{ccc}{\eta_{ij_1}\vec{x}_{ij_1} - \eta_{ik}\vec{x}_{ik} &\eta_{ij_2}\vec{x}_{ij_2} - \eta_{ik}\vec{x}_{ik}&\eta_{ij_3}\vec{x}_{ij_3} - \eta_{ik}\vec{x}_{ik}} = \\
    \det(\M{K}_m^{-1})\, \det \mat{ccc}{\eta_{mj_1}\vec{x}_{mj_1} - \eta_{mk}\vec{x}_{mk} &\eta_{mj_2}\vec{x}_{mj_2} - \eta_{mk}\vec{x}_{mk}&\eta_{mj_3}\vec{x}_{mj_3} - \eta_{mk}\vec{x}_{mk}} 
\end{multline*}
The above equations can be concisely rewritten as
\begin{gather}
    \eta_{ij}^2\,\vec{x}_{ij}^\top\,\M{K}_i^{-\top}\M{K}_i^{-1}\vec{x}_{ij} - 2\,\eta_{ij}\,\eta_{ik}\,\vec{x}_{ij}^\top\,\M{K}_i^{-\top}\M{K}_i^{-1}\vec{x}_{ik} + \eta_{ik}^2\,\vec{x}_{ik}^\top\,\M{K}_i^{-\top}\M{K}_i^{-1}\vec{x}_{ik} = \hspace*{6cm} \nonumber \\
    \hspace*{4cm}
    \eta_{mj}^2\,\vec{x}_{mj}^\top\,\M{K}_m^{-\top}\M{K}_m^{-1}\vec{x}_{mj} - 2\,\eta_{mj}\,\eta_{mk}\,\vec{x}_{mj}^\top\,\M{K}_m^{-\top}\M{K}_m^{-1}\vec{x}_{mk} + \eta_{mk}^2\,\vec{x}_{mk}^\top\,\M{K}_m^{-\top}\M{K}_m^{-1}\vec{x}_{mk} \label{eq:e^2xKiTKix-...=e^2xKiTKix}
    \\[1ex]
    \det(\M{K}_i^{-1})\,\det \mat{ccc}{\eta_{ij_1}\vec{x}_{ij_1} - \eta_{ik}\vec{x}_{ik} &\eta_{ij_2}\vec{x}_{ij_2} - \eta_{ik}\vec{x}_{ik}&\eta_{ij_3}\vec{x}_{ij_3} - \eta_{ik}\vec{x}_{ik}} = \hspace*{6cm} \nonumber \\
    \hspace*{4cm}
    \det(\M{K}_m^{-1})\, \det \mat{ccc}{\eta_{mj_1}\vec{x}_{mj_1} - \eta_{mk}\vec{x}_{mk} &\eta_{mj_2}\vec{x}_{mj_2} - \eta_{mk}\vec{x}_{mk}&\eta_{mj_3}\vec{x}_{mj_3} - \eta_{mk}\vec{x}_{mk}} 
    \label{eq:det(K_i^-1)...=det(K_m^-1)}
    \\[1ex]
    h(\eta_{ij})=0,\ f_l(\M{K}_i) = 0,\ \det{\M{K}_i} \neq 0,\ \eta_{ij} \neq 0 
    \label{eq:hf|k|e=/=0}\\[1ex]
    i, m=1,\ldots,M,\ j_1,j_2,j_3=1,\ldots,N,\ k=1,\ldots,N,\ l=1,\ldots,L  \label{i,j123,k.l=...}
\end{gather}

Now, we can ask whether polynomials defined by~ \cref{eq:e^2xKiTKix-...=e^2xKiTKix,eq:det(K_i^-1)...=det(K_m^-1),eq:hf|k|e=/=0} generate the elimination ideal obtained by intersecting the ideal generated by~\cref{eq:zx=K[R|-RC][X;;],eq:K=[...oool]} with $\QQ[\eta_{ij},\M{K}_i]$. Before addressing this question in general, we can observe that it is true for two calibrated views.

\begin{observation}
    Assuming $M=2$ calibrated cameras observing  $N=5$ points   with projection equations
    \begin{equation*}
    \begin{split}
    F = \left\{
    \eta_{1j}\,\vec{x}_{1j} -
    \mat{cc}{\M{I} & 0}\,\mat{c}{\vec{X}_j\\1},\
    \eta_{2j}\,\vec{x}_{2j} - 
    \M{R}_2\mat{cc}{\M{I} &-\vec{C}_2}\,\mat{c}{\vec{X}_j\\1}, \
    \M{R}_2^\top\M{R}_2-\M{I},\ \det{\M{R}_2}-1,\ 
    \zeta_{ij}\eta_{ij}-1, \eta_{11}-1
    \ \middle| \right. \quad \\ 
    \left. \phantom{\mat{c}{\vec{X}_j\\1}} 
    \zeta_{ij}, \eta_{ij} \in \CC,\ 
    x_{ij}, X_j, C_2 \in \CC^3,\  
    R_2 \in \CC^{3 \times 3},\ 
    i=1,\ldots,2,\ j=1,\ldots,5 \right\}
    \end{split}
    \end{equation*}
    we consider ...
\end{observation}
\begin{proof}
    Consider ideal $I = \langle F \rangle$ in ring $\CC[X_j,C_2,R_2,\zeta_{ij},\eta_{ij},x_{ij}]$. Clearly, 
    \begin{eqnarray*}
        I &=& \left\langle
        \vec{X}_j-\eta_{1j}\,\vec{x}_{1j},\
        \vec{X}_j-\vec{C}_2-\eta_{2j}\,\M{R}_2^\top\vec{x}_{2j},\
        \M{R}_2^\top\M{R}_2-\M{I},\ \det{\M{R}_2}-1,\ 
        \zeta_{ij}\eta_{ij}-1, \eta_{11}-1
        \right\rangle \\
        &=&\left\langle
        \vec{X}_j-\eta_{1j}\,\vec{x}_{1j},\
        \vec{C}_2+\eta_{2j}\,\M{R}_2^\top\vec{x}_{2j}-\eta_{1j}\,\vec{x}_{1j},\
        \M{R}_2^\top\M{R}_2-\M{I},\ \det{\M{R}_2}-1,\ 
        \zeta_{ij}\eta_{ij}-1, \eta_{11}-1
        \right\rangle\\
        &=&\langle
        \vec{X}_j-\eta_{1j}\,\vec{x}_{1j},\
        \vec{C}_2+\eta_{21}\,\M{R}_2^\top\vec{x}_{21}-\eta_{11}\,\vec{x}_{11},\ 
        \M{R}_2^\top\M{R}_2-\M{I},\ \det{\M{R}_2}-1,\\
        & & \phantom{\langle}
        \M{R}_2^\top(\eta_{2j}\,\vec{x}_{2j}-\eta_{21}\,\vec{x}_{21})
        -\eta_{1j}\,\vec{x}_{1j}+\eta_{11}\,\vec{x}_{11},\
        \zeta_{ij}\eta_{ij}-1, \eta_{11}-1
        \rangle
    \end{eqnarray*}
    Now, when combining $\M{R}_2^\top(\eta_{2j}\,\vec{x}_{2j}-\eta_{21}\,\vec{x}_{21})
        -\eta_{1j}\,\vec{x}_{1j}+\eta_{11}\,\vec{x}_{11}$ and $\M{R}_2^\top\M{R}_2-\M{I}$, we get for $j=1,\ldots,5$
    \begin{multline*}
    (\M{R}_2^\top(\eta_{2j}\,\vec{x}_{2j}-\eta_{21}\,\vec{x}_{21})+\eta_{1j}\,\vec{x}_{1j}-\eta_{11}\,\vec{x}_{11})^\top \M{R}_2^\top((\eta_{2j}\,\vec{x}_{2j}-\eta_{21}\,\vec{x}_{21})-\eta_{1j}\,\vec{x}_{1j}+\eta_{11}\,\vec{x}_{11}) \\
    - ((\eta_{2j}\,\vec{x}_{2j}-\eta_{21}\,\vec{x}_{21})+\eta_{1j}\,\vec{x}_{1j}-\eta_{11}\,\vec{x}_{11})^\top (R_2^T R_2-I) ((\eta_{2j}\,\vec{x}_{2j}-\eta_{21}\,\vec{x}_{21})-\eta_{1j}\,\vec{x}_{1j}+\eta_{11}\,\vec{x}_{11})
    = \\
    (\eta_{2j}\,\vec{x}_{2j}-\eta_{21}\,\vec{x}_{21})^\top(\eta_{2j}\,\vec{x}_{2j}-\eta_{21}\,\vec{x}_{21}) - 
    (\eta_{1j}\,\vec{x}_{1j}-\eta_{11}\,\vec{x}_{11})^\top(\eta_{1j}\,\vec{x}_{1j}-\eta_{11}\,\vec{x}_{11})
    \end{multline*}    
    Similarly, when combining $\M{R}_2^\top(\eta_{2j}\,\vec{x}_{2j}-\eta_{21}\,\vec{x}_{21})
        -\eta_{1j}\,\vec{x}_{1j}+\eta_{11}\,\vec{x}_{11}$ and $\det{\M{R}_2}-1$ we get for $i_1,i_2,i_3,k \in 1,\ldots,5$
    \begin{multline*}
    \det \mat{ccc}{
    \eta_{2j_1}\vec{x}_{2j_1} - \eta_{2k}\vec{x}_{2k} &
    \eta_{2j_2}\vec{x}_{2j_2} - \eta_{2k}\vec{x}_{2k} &
    \eta_{2j_3}\vec{x}_{2j_3} - \eta_{2k}\vec{x}_{2k}} - \\
    \det \mat{ccc}{
    \eta_{mj_1}\vec{x}_{1j_1} - \eta_{1k}\vec{x}_{1k} & 
    \eta_{mj_2}\vec{x}_{1j_2} - \eta_{1k}\vec{x}_{1k} &
    \eta_{mj_3}\vec{x}_{1j_3} - \eta_{1k}\vec{x}_{1k}} 
    \end{multline*}
Thus,
\begin{eqnarray}
  I &=&\langle
        \vec{X}_j-\eta_{1j}\,\vec{x}_{1j},
        \label{eq:X-ex} \\ & & \phantom{\langle}
        \vec{C}_2+\eta_{21}\,\M{R}_2^\top\vec{x}_{21}-\eta_{11}\,\vec{x}_{11},
        \label{eq:C-eRx} \\ & & \phantom{\langle}
        \det{\M{R}_2}-1,\ \M{R}_2^\top\M{R}_2-\M{I},
        \label{eq:detR} \\ & & \phantom{\langle}
        \M{R}_2^\top(\eta_{2j}\,\vec{x}_{2j}-\eta_{21}\,\vec{x}_{21})
        -\eta_{1j}\,\vec{x}_{1j}+\eta_{11}\,\vec{x}_{11},
        \label{eq:R(ex)} \\ & & \phantom{\langle}
        \zeta_{ij}\eta_{ij}-1,
        \label{eq:ze-o} \\ & & \phantom{\langle}
        (\eta_{2j}\,\vec{x}_{2j}-\eta_{21}\,\vec{x}_{21})^\top(\eta_{2j}\,\vec{x}_{2j}-\eta_{21}\,\vec{x}_{21}) - 
        (\eta_{1j}\,\vec{x}_{1j}-\eta_{11}\,\vec{x}_{11})^\top(\eta_{1j}\,\vec{x}_{1j}-\eta_{11}\,\vec{x}_{11}), 
        \label{eq:ex-d} \\ & & \phantom{\langle}
        \det \mat{ccc}{
        \eta_{2j_1}\vec{x}_{2j_1} - \eta_{2k}\vec{x}_{2k} &
        \eta_{2j_2}\vec{x}_{2j_2} - \eta_{2k}\vec{x}_{2k} &
        \eta_{2j_3}\vec{x}_{2j_3} - \eta_{2k}\vec{x}_{2k}} - 
        \label{eq:ex-o} \\ & & \phantom{\langle}
        \hspace*{4cm} \det \mat{ccc}{
        \eta_{1j_1}\vec{x}_{1j_1} - \eta_{1k}\vec{x}_{1k} & 
        \eta_{1j_2}\vec{x}_{1j_2} - \eta_{1k}\vec{x}_{1k} &
        \eta_{1j_3}\vec{x}_{1j_3} - \eta_{1k}\vec{x}_{1k}}, 
        \nonumber \\ & & \phantom{\langle}
        \eta_{11}-1 \label{eq:e-o}
        \rangle
    \end{eqnarray}
    Let us find generators of the elimination ideal $J = I \cap \CC[\zeta_{ij},\eta_{ij},x_{ij}]$. We consider an elimination monomial order~\cite{Cox-IVA-2015} with blocks of variable ordered as $X_j > C_2 > R_2 > \zeta_{ij} > \eta_{ij} > x_{ij}$. Let us introduce the set $S(F,G) = \{S(f,g) \mid f \in F, g \in G\}$ of S-polynomials constructed from sets of polynomials $F, G$.
    
    \begin{tabular}{|c||c|c|c|c|c|c|c|c|}
         \hline 
         $\overline{S(\downarrow,\rightarrow)}$&(\ref{eq:X-ex})&(\ref{eq:C-eRx})&(\ref{eq:detR})&(\ref{eq:R(ex)})&(\ref{eq:ze-o})&(\ref{eq:ex-d})&(\ref{eq:ex-o})&(\ref{eq:e-o}) \\ \hline \hline
         (\ref{eq:X-ex})  & 0 & 0 & 0 & 0 &     0 &     0 &      0 & 0    \\ \hline
         (\ref{eq:C-eRx}) & 0 & 0 & 0 & 0 &     0 &     0 &      0 & 0    \\ \hline
         (\ref{eq:detR})  & 0 & 0 &$I_{R_2}$& & 0 &     0 &      0 & 0    \\ \hline
         (\ref{eq:R(ex)}) & 0 & 0 &   &   &       &       &        &      \\ \hline
         (\ref{eq:ze-o})  & 0 & 0 & 0 &   &$\in J$&$\in J$&$\in J$&$\in J$\\ \hline
         (\ref{eq:ex-d})  & 0 & 0 & 0 &   &$\in J$&$\in J$&$\in J$&$\in J$\\ \hline
         (\ref{eq:ex-o})  & 0 & 0 & 0 &   &$\in J$&$\in J$&$\in J$&$\in J$\\ \hline
         (\ref{eq:e-o})   & 0 & 0 & 0 &   &$\in J$&$\in J$&$\in J$&$\in J$\\ \hline
    \end{tabular}

    We notice that S-polynomials constructed from generators~(\ref{eq:X-ex}) and~(\ref{eq:C-eRx}) will all reduce to 0 since the leading terms $X_j$ and $C_2$ are co-prime to all other leading monomials.  We see that $S((\ref{eq:detR}),(\ref{eq:detR}))$ will be in $I_{R_2} = I \cap \CC[R_2]$

    $h = \sum_{f,g \in F \cup G} f g$
\end{proof}

\begin{proof}
The following Macaulay2 code shows that the observation is true for generic situations:     
{\tiny
\begin{verbatim}
restart
needs "top.m2";
pN = 5; -- the number of points
cN = 2; -- the number of cameras
FF = ZZ/30011 
-- atocalibration problem setup
Rg = FF[c_(0,0)..c_(cN-2,2),t_(0,0)..t_(cN-2,2),x_(0,0)..x_(2,pN-1),e,z_(0,0)..z_(cN-1,pN-1),g_(0,0)..g_(cN-1,pN-1)] 
K = mat {{1_Rg,0,0},{0,1,0},{0,0,1}}
iK = K 
K0 = mat {{1_FF,0,0},{0,1,0},{0,0,1}}
-- simulated scene 
iK0 = inv(K0)
X0 = a2h(mat(randomMutableMatrix(3,pN,0.0,100)) + 100*ones(3,pN)) -- random points infront of the camera
X  = a2h(mat(genericMatrix(Rg,x_(0,0),3,pN)))  -- point variables
T0 = {zeros(3,1)} | apply(cN-1,i->random(FF^3,FF^1)) -- random translations 
T  = {zeros(3,1)} | apply(cN-1,i->mat(genericMatrix(Rg,t_(i,0),3,1)))  -- point variables
R0 = {E3} | apply(cN-1,i->c2R(mat(random(FF^3,FF^1)))) -- random rotations
R  = {E3} | apply(cN-1,i->C2R(mat(genericMatrix(Rg,c_(i,0),3,1)))) -- unknown rotation matrices
P0 = apply(cN,i->K0*(R0#i)*(sub(E3,FF) | -1_FF*(T0#i))) 
P  = apply(cN,i->K*(R#i)*(sub(E3,FF) | -1_FF*(T#i))) -- Projection matrices
x0 = apply(P0,P->sub(P*X0,Rg)) -- image projections
G = apply(cN,i->genericMatrix(Rg,g_(i,0),pN,1)) -- depths
-- setup equations
d = {1} | toList apply(1..cN-1,i->(first ent gens first decompose ideal det(R#i))) -- the divisors for R matrices to make det=1
eq = apply(cN,i->(x0#i)*dia(G#i)*(d#i)-(P#i)*X) -- projecton equations
ds = flatten for i from 0 to cN-1 list (for j from 0 to pN-1 list g_(i,j)*z_(i,j)-1); -- avoid g_(i,j)=0
I  = ideal eq + -- projection equations
     ideal(g_(0,0)-1) + -- dehomogenization
     ideal(ds) + -- no g is zero
     ideal(e*d#1-1); -- remove a 9 dim spurious component
dim I, degree I -- cN=2, pN=5: (0,20) 
B = ent gens gb I; -- #B = 213
-- Compare to the "depth formulation"
ip = subsets(pN,2) -- point pairs
de = flatten apply(ip,i->apply(subsets(cN,2),j->N2(G#(j#0)_(i#0,0)*iK*x0#(j#0)_{i#0}-G#(j#0)_(i#1,0)*iK*(x0#(j#0))_{i#1})
                                               -N2(G#(j#1)_(i#0,0)*iK*x0#(j#1)_{i#0}-G#(j#1)_(i#1,0)*iK*(x0#(j#1))_{i#1})));
-- GB of the elimination ideal in z, g
load "FGLM.m2"
Rg2 = FF[c_(0,0)..c_(cN-2,2),t_(0,0)..t_(cN-2,2),x_(0,0)..x_(2,pN-1), e, z_(0,0)..z_(cN-1,pN-1), g_(0,0)..g_(cN-1,pN-1), MonomialOrder=>{21,1,10,10}]
Io = ideal gens fglm(I,Rg2); -- Io == I in a different ordering
Id = sub(ideal(de),Rg2) + 
     sub(ideal(g_(0,0)-1),Rg2) + 
     sub(ideal(ds),Rg2); -- the depths formulation ideal
Rg3 = FF[z_(0,0)..z_(cN-1,pN-1), g_(0,0)..g_(cN-1,pN-1),MonomialOrder=>{10,10}]
Id = sub(Id,Rg3)
Ie = sub(ideal selectInSubring(2,gens Io),Rg3); -- elimination ideal in g's, i.e. Ie = eliminate(Io,(ent vars Rg2)_{0..21});
-- Constraints on g's to ensure det(R)=+1
n = apply(cN,i->gens minors(3,(x0#i)*dia(G#i)-(x0#i_{0}*(G#i)_(0,0)*ones(1,cdim(x0#i)))))
D = apply(subsets(cN,2),i->n#(i#0)-n#(i#1))
K = Id + sub(ideal D, Rg3)
degree K -- 20
K == Ie -- true => depth equations + det R = 1 constraints generate the elimination ideal from the original projection equations with g's non-zero 

o50 = true
\end{verbatim}}
\end{proof}

\newpage
Next, we can introduce a polynomial map $\varphi\colon \eta_{ij} \mapsto \eta_{ij},\ \M{O}_i \mapsto \M{K}_i^{-\top}\M{K}_i^{-1}$ that pulls Equations~\ref{eq:e^2xKiTKix-...=e^2xKiTKix}, \ref{eq:hf|k|e=/=0} back into
\begin{align}
     \eta_{ij}^2\,\vec{x}_{ij}^\top\,\M{O}_i\,\vec{x}_{ij} - 2\,\eta_{ij}\,\eta_{ik}\,\vec{x}_{ij}^\top\,\M{O}_i\,\vec{x}_{ik} + \eta_{ik}^2\,\vec{x}_{ik}^\top\,\M{O}_i\,\vec{x}_{ik} =
     \eta_{mj}^2\,\vec{x}_{mj}^\top\,\M{O}_m\,\vec{x}_{mj} - 2\,\eta_{mj}\,\eta_{mk}\,\vec{x}_{mj}^\top\,\M{O}_m\,\vec{x}_{mk} + \eta_{mk}^2\,\vec{x}_{mk}^\top\,\M{O}_m\,\vec{x}_{mk} \nonumber \\
    h(\eta_{ij})=0,\ \eta_{ij} \neq 0 \nonumber \\
    \M{O}_i^\top = \M{O}_i,\ \det{\M{O}_i} \neq 0,\ \det{\M{O}_i(1:2,1:2)}^2=\det{\M{O}_i},\ e_l(\M{O}_i) = 0 
    \nonumber \\
    i=1,\ldots,M,\ m=i+1,\ldots,M,\ j=1,\ldots,N,\ k=j+1,\ldots,N,\ l=1,\ldots,L   
\end{align}
Map $\varphi$ maps four solutions for each $\M{K}_i$ into one solution for each $\M{O}_i$ since there are four ways how to decompose a symmetric $\M{O}_i$ into an upper triangular $\M{K}_i$ with $\M{K}_i(3,3)=1$. Equations $\det{\M{O}_i(1:2,1:2)}^2=\det{\M{O}_i}$ follow from $\M{K}_i(3,3)=1$ and equations $e_l(\M{O}_i) = 0$ follow from $f_l(\M{K}_i) = 0$.

Now, let us consider non-equalities $\eta_{ij} \neq 0$ and $\det{\M{O}_i} \neq 0$. The standard way of imposing non-equality constraints is to introduce additional variables $z_{ij}$, $z_{i}$ and add equations $z_{ij}\eta_{ij}-1=0$, $z_i\det{\M{O}_i}-1=0$~\cite{Cox-IVA-2015}. The disadvantage of this approach is that it introduces additional variables and makes the computations more expensive. Here, we will use a particular rational map, which appeared in~\cite{DBLP:journals/jmiv/QuanTM06}, to remove the spurious positive dimensional components of ${\mathcal V}$. 

We construct rational map $\psi\colon \M{O}_i \mapsto \M{O}_i,\ $

\newpage
\section{Graph Isomorphism}

\noindent
\textbf{Notation.} Let us denote a generic 3D point as $x_p$. The Euclidean inter-point distance between two arbitrary points $x_p$ and $x_q$ is denoted as $\delta_{pq}$. The number of views is denoted as $N$, whereas the number of 3D points is denoted as $P$. The projective depth of a generic 3D point $x_p$ with respect to a generic view $i$ is denoted as $\lambda_{ip}$. We assume that all views are captured by one or more cameras with shared intrinsic parameters. The set of intrinsic camera parameters is denoted as $\mathcal{K} = \{ k_i \}_{i=1}^5$.

\medskip
\noindent
\textbf{Effectiveness of the cosine-rule formulation for 3V4P.} In the traditional 3V4P (Scranton) formulation, the system of cosine-rule polynomials is derived by eliminating the inter-point distances $\delta_{pq}$ between generic points $x_{p}$ and $x_q$, as observed by two view pairs $(i,j)$ and $(i,k)$ in the system. This formulation is particularly advantageous given a set of $4$ distinct 3D points. In such a scenario, the $4$ points inherently have $6$ Euclidean invariants, computed as the product of three spatial dimensions $3 \times 4$ d.o.f. minus 6 d.o.f. of the Euclidean transformation between the generic view pair $(i, j)$. Remarkably, this count of invariants is equal to the number of inter-point distances $\delta_{pq}$. From a geometric perspective, the pairwise distances $\delta_{pq}$ can be interpreted as the edge lengths of a tetrahedron. Equivalently, the $4$ 3D points can be seen as the vertices $V$ of a graph $G = (V, E)$ for which the length of each edge $e \in E$ corresponds to one of the inter-point distances $\delta_{pq}$.

\smallskip
Given $N=3$ images of $P=4$ points, we obtain a system of $\binom{P}{2}N = 6N = 18$ homogeneous polynomials $f(\lambda_{ip}, \lambda_{iq}, \delta_{pq}) = 0, \; i = 1, \ldots, N, \; p < q = 1, \ldots, 4$. By eliminating the inter-point distances $\delta_{pq}$, we obtain $12$ homogeneous polynomials in $12$ unknowns $\lambda_{ip}$. After de-homogenization $(\lambda_{11} = 1)$, we obtain $11$ in-homogeneous unknowns $\lambda_{ip}$, meaning that the problem is over-constrained, given the $12$ independent constraints available. Existing solvers for the 3V4P (Scranton) problem introduce a slack variable $l$ to balance the number of unknowns and constraints. Alternatively, the additional constraint can be leveraged to overcome the assumption of a calibrated camera in 3V4P and, instead, assume that one of the $\{ k_i \}_{i=1}^{5}$ intrinsic camera parameters is unknown. In such a scenario, the resulting problem is balanced in the number of constraints and unknowns, with $12$ polynomials in 11 in-homogeneous unknowns $\lambda_{ip}$ plus an unknown calibration parameter $k_i \in \{ k_i \}_{i=1}^{5}$. For instance, the aforementioned formulation would allow to solve for the focal length of the camera, assuming $f = f_x = f_y$.

\medskip
\noindent
\textbf{Extension to $\mathbf{P > 4}$.}
In the eventuality that more than one camera parameter $k_i \in \{ k_i \}_{i=1}^{5}$ is unknown, it is evident that joint Euclidean reconstruction and camera calibration cannot be achieved for $N=3$ views and $P=4$ points. Alas, when the number of unknown intrinsic parameters $K > 1$, the problem is under-constrained and, thus, an additional point is required to constrain the introduced unknown camera parameters. The extension of the proposed inter-point distance formulation is, however, not trivial for $N=3$ views and $P > 4$ points.

The inter-point distance formulation is not as effective when considering $P > 4$ 3D points. In fact, for $P > 4$, the inter-point distances $\delta_{pq}$ and their corresponding cosine-rule polynomials are not all independent. Precisely, for $P$ points in 3D space, the count of Euclidean invariants for the set of points can be expressed as $3 \times P$. However, for $P > 4$, the number of inter-point distances $\delta_{pq}$ surpasses the count of Euclidean invariants of the 3D points, as expressed by the inequality:
\begin{equation}
    3 \times P < \binom{P}{2} \; , \; \; P > 4.
\end{equation}
As the number of derived cosine-rule polynomials surpasses the number of degrees of freedom in the system, it is evident that the inter-point distances $\delta_{pq}$ are not all independent. Thus, to solve the polynomial system for a finite set of solutions, it is imperative to selectively omit cosine-rule polynomials so that the resulting system consists only of constraints that are all independent. 

For instance, for $P = 5$, the number of Euclidean invariants is $3 \times 5 = 15$ d.o.f., and, when subtracting the $6$ d.o.f. attributed to the Euclidean transformation, we obtain $9$ independent Euclidean invariants. In contrast, the number of inter-point distances, determined by $\binom{5}{2} = 10$, exceeds the count of independent Euclidean invariants.

\medskip
\noindent
\textbf{Relationship to Minimally Rigid Graphs.} Given $P$ 3D points observed by a view pair $(i,j)$, the $P$ points can be seen as the set of vertices $V$ within a graph $G = (V, E)$ in $\mathbb{R}^3$. In such a graph, the edges $E$ correspond to the inter-point distances $\delta_{pq}$. A relationship can thus be established between the independence of constraints in the polynomial system and the theory of minimally rigid graphs. Building upon the derivations in \cite{}, we can assert that a graph $G$ representing $P$ 3D points is minimally rigid when it contains precisely $3|P| - \binom{d + 1}{2}$ edges, with $d=3$. This condition holds true provided that $|V| \geq d + 1 = 4$, and is always satisfied in our case as the Euclidean reconstruction of points in 3D space necessitates a minimum of $4$ points \cite{}. In essence, the expression $3|V| - \binom{4}{2} = 3 \times N - 6$ is equivalent to the number of Euclidean invariants in a system of $P$ points.

\subsection{Euclidean Reconstruction from 3 Views with Multiple Unknown Calibration Parameters}
We present the challenges of Euclidean reconstruction and camera calibration when $P$ 3D points are visible from $N=3$ views and the number of unknown camera parameters $K \geq 2$. Following the derivations for the 3V4P problem, for $N=3$ and $P$ 3D points, we obtain a system of $\binom{P}{2} \times N = \binom{P}{2} \times 3$ cosine-rule polynomials in $N \times P = 3 \times P$ unknowns $\lambda_{ip}$ and $\binom{P}{2}$ unknowns $\delta_{ij}$. By eliminating the inter-point distances $\delta_{ij}$, we obtain $\binom{P}{2} \times (N - 1) = \binom{P}{2} \times 2$ polynomials in $N \times P - 1$ inhomogeneous unknowns $\lambda_{ip}$. 


\medskip
\noindent
\textbf{Multi-Graph Encoding.} For $N \geq 3$, a polynomial system of cosine-rule constraints can be encoded as a labelled multi-graph $\mathcal{G} = (V, E)$, where each vertex $p \in V$ corresponds to a 3D point $x_p$, and, thus, $|V| = P$. Two vertices $p,q \in V$ are connected by an edge $(p,q) \in E$ labelled $(i,j)$ if the cosine-rule constraint involving the point pair $(x_p, x_q)$ and views $(i,j)$ appears in the encoded polynomial system of $U$ equations, where $U$ is the number of unknowns in the system, i.e., the set of projective depths $\lambda_{ip}$ and a subset $\mathcal{K}' \subseteq \mathcal{K}$ of intrinsic camera parameters, with $K = |\mathcal{K}'|$. The multi-graph $\mathcal{G}$ satisfies the property that, $\forall (p,q) \in E$, the degree of the multi-edge $d_{pq} \leq N - 1$, as independent constraints can only be derived from $N - 1$ view pairs.

\medskip
\noindent
\textbf{Balancing Unknowns and Equations.} The number $K$ of unknown intrinsic camera parameters $\mathcal{K}' \subseteq \mathcal{K}$ directly reflects into the number of unknowns in which the system of cosine-rule polynomials can be expressed. Therefore, to balance the number of equations and the number of unknowns in the system, a subset of $\binom{P}{2} - U$ equations must be removed from the system consisting of all available cosine-rule polynomials, i.e., the system corresponding to the complete multi-graph $\mathcal{G}$ where each multi-edge $(p,q) \in E$ has multiplicity $d_{pq} = N - 1$, i.e., its maximum allowed multiplicity.

\medskip
\noindent
\textbf{The Problem of Removing Equations.} 
Two problems may arise by removing equations from the polynomial system consisting of all available cosine-rule polynomials. First, by omitting a subset of equations, the system may still consist of constraints that are not all independent. For instance, this problem emerges in the 3V5P problem with $K=|\mathcal{K}'|=4$ when removing $2$ constraints involving the first view pair, while retaining all $10$ constraints from the second view pair, which are not all independent. Second, removing equations may result in having a set of all independent polynomials that, however, do not constrain all Euclidean invariants and intrinsic calibration parameters in the system, as we shall discuss in later sections. In such a case, the set of solutions to the system is not finite, as one or more degrees of freedom remain unconstrained. For $N=2$ views, identifying which constraints to remove in order to balance the number of unknowns in the system is straightforward, as the system can be expressed as a simple graph for which it is trivial to determine its minimal rigidity. However, extending the problem to $N \geq 3$ is not straightforward, as the minimal rigidity of the multi-graph encoding our polynomial system is hard to define and determine. As already discussed, the system can be encoded as a multi-graph with edges of maximum multiplicity $d_{pq} = N - 1$. To the best of our knowledge, neither a theoretical definition nor proof for minimal rigidity of multi-graphs have not been devised and it has never been proven whether a direct relationship between our problem and minimal rigidity of multi-graphs exists. \tim{Not sure if ``minimally rigid multigraph" even makes much sense---two edges between the same vertices should realize the same distance, so I think this would just reduce to the case of graphs.}
Also, as we shall demonstrate later, in general, the rigidity of individual simple graphs derived by considering only one view pair $(i,j)$ at a time, i.e., a graph with the same vertices $V$ and edges $(p,q) \in E$ only labelled as $(i,j)$, is not directly related to the ability to solve the polynomial system.

\medskip
\noindent
\textbf{Taxonomy of Minimal Problems in 3V?P.}
Given the considerations so far, we identify the need to provide a complete taxonomy of all possible balanced problems for $N=3$ views and $P \geq 4$ 3D points to exhaustively identify which multi-graphs correspond to polynomial systems that can be solved for a finite number of solutions with the goal of achieving Euclidean reconstruction and, when necessary, camera calibration from a minimal set of points. 

\medskip
\noindent
\textbf{3V4P} supports Euclidean reconstruction, as well as partial calibration, when $K \in \{ 0, 1 \}$ and is characterized by $12$ cosine-rule polynomials, which are all independent.
\begin{itemize}
    \item $K = 0$, i.e., $11$ unknowns \\
    The problem reduces to the classical, i.e., "Scranton", formulation. In \cite{}, the problem is balanced by introducing a slack variable, which acts as an additional degree of freedom in the system. Alternatively, the problem can be balanced by omitting $1$ equation from the system, leading to $12$ unique ways to remove equations.
    \item $K = 1$, i.e., $12$ unknowns \\
    This is a minimal problem that is perfectly balanced out of the box.
\end{itemize}

\medskip
\noindent
\textbf{3V5P} supports Euclidean reconstruction, as well as partial calibration, when $K \in \{ 2, 3, 4 \}$ and is characterized by $20$ cosine-rule polynomials, which are \emph{not} all independent.
\begin{itemize}
    \item $K = 2$, i.e., $16$ unknowns \\
    The problem can be balanced by omitting $4$ equations from the system, leading to $4845$ unique multi-graphs representing polynomial system configurations.
    \item $K = 3$, i.e., $17$ unknowns \\
    The problem can be balanced by omitting $3$ equations from the system, leading to $1140$ unique multi-graphs representing polynomial system configurations.
    \item $K = 4$, i.e., $18$ unknowns \\
    The problem can be balanced by omitting $2$ equations from the system, leading to $190$ unique multi-graphs representing polynomial system configurations.
\end{itemize}

\medskip
\noindent
\textbf{3V6P} supports Euclidean reconstruction and full camera calibration when $K = 5$ and is characterized by $30$ cosine-rule polynomials, which are \emph{not} all independent.
\begin{itemize}
    \item $K = 6$, i.e., $22$ unknowns \\
    The problem can be balanced by omitting $8$ equations from the system, leading to $5852925$ unique multi-graphs representing polynomial system configurations.
\end{itemize}

\medskip
\noindent
\textbf{Which configurations can be solved?} Previously, we discussed how it is hard to determine a priori whether a given multi-graph corresponds to a polynomial system that can be solved for a finite number of solutions. The number of configurations available is substantial, making it difficult to provide a taxonomy of which balanced minimal problems can actually be solved for a finite set of solutions and the number of solutions that this would entail. In addition, depending on which subset of the calibration parameters $\mathcal{K}' \subseteq \mathcal{K}$ is unknown, the number of solutions to the system may vary.

\medskip
\noindent
\textbf{Permutation of 3D points and View Pairs.}
We simplify the problem of identifying which unique ways of removing constraints, i.e., multi-graphs, produce polynomial systems that can be solved by introducing equivalence classes between labelled multi-graphs. First, let us discuss why we can introduce equivalence classes. Given a multi-graph $G = (V,E)$, each vertex $p \in V$ corresponds to a generic 3D point. Let us now provide an index to the vertices in the multi-graph as $V = \{ v_1, v_2, \ldots, v_P \}$, where $P$ is the number of 3D points in the system. A permutation of the indices assigned to the vertices $V$ in the multi-graph does not alter the overall structure of the 3D points in space, as their coordinates remain unchanged, but determines a change in the points that are involved in the cosine-rule polynomials present in the polynomial system. We claim that a permutation of the indices of the vertices does not alter the solvability of the system nor its number of complex solutions. This is due to the fact that 3D points are assumed to have generic coordinates, and, thus, are projected to also generic 2D points on the image plane of a given view. As the image points are generic and constitute the only parameters of the polynomial system, we expect to obtain the same number of solutions after a vertex permutation. The same principles apply to the permutation of image pairs, as changing the order in which view pairs are considered does not have an impact on the solvability or number of complex solutions in the system.

\medskip
\noindent
\textbf{Determining Equivalence Classes.} We assign a label to each vertex in the multi-graph depending on the multiplicity of its outgoing edges. Specifically, we design the labels of vertices as the set of multiplicities of the outgoing edges of a given vertex in the multi-graph. Formally, given a vertex $v \in V$, the set of outgoing edges $E_v = \{ (p,q) : p = v, (p,q) \in E \} $ can be used to derive the label of $v$ as $L'_v = \{ d_{pq}: (p,q) \in E_v \}$, where $d_{pq}$ is the multiplicity of the multi-edge $(p,q)$. The final label $L_v$ is obtained by ordering the elements in the label $L'_v$. For visualization purposes, we can assign the each multiplicity value $d_{pq}$ to a color value, also considering that the maximum multiplicity of an edge $d_{pq} = N - 1 = 2$. In our illustrations:
\begin{itemize}
    \item $d_{pq}=0$: white
    \item $d_{pq}=1$: gray
    \item $d_{pq}=2$: black
\end{itemize}



\subsection{Legacy}
From previous derivations we know that:
\begin{enumerate}
    \item 3V4P, when $K \in \{ 0, 1 \}$. When $K = 0$, we are dealing with the "so-called" Scranton problem, which has been solved \cite{}. When $K = 1$, the problem is perfectly balanced with $12$ unknowns in $12$ equations.
    \item 3V5P, when $K \in \{ 2, 3, 4 \}$, as the number of independent cosine-rule polynomials that can be derived from $P = 5$ points is $18$, with $14$ in-homogeneous unknowns in $\lambda_{ip}$. For $K = 2$, then we have a total of $14 + 2 = 16$ unknowns and $20$ equations, meaning that we have $\binom{20}{16} = \binom{20}{4} = 4085$ different ways of removing constraints from our system, i.e., edges from our multi-graph, leading to $4085$ different multi-graphs with $|V| = P = 5$ and $|E| = 14 + 2 = 16$ with $\forall (p,q) \in E, d_{pq} \leq N - 1$. Following the same rationale for $K = 3$, we obtain $1140$ unique multi-graphs, whereas for $K = 2$, we obtain 190 unique multi-graphs.
    \item 3V6P, when $K = 5$. In this case, the number of unknowns $3 \times P - 1 + U = 3 \times 6 - 1 + 5 = 22$, whereas the number of equations is $\binom{6}{2} \times (N - 1) = 30$. This leads to $\binom{30}{22} = 5852925$ unique multi-graphs.
\end{enumerate}

previous calculation that for each view pair, , we know that the polynomial system built from constraints of individual view pairs $i < j, 1, \ldots, 3$ will contain at least $2$ constraints that are dependent on others, one for each view pair. This leaves with $18$ independent constraints and $14$ in-homogeneous unknowns $\lambda_{ip}$, meaning that there is space for calibration of up to $4$ unknown intrinsic camera parameters. 

Depending on the number of unknown calibration parameters $K \in \{ 2, 3, 4 \}$, we will need to relax our polynomial system by removing a certain number of equations. In general, given $K$ unknown calibration parameters, we need to remove from the system XX constraints, meaning that there is a total of $\binom{\binom{P}{2} \times (N - 1)}{K + N \times P - 1}$ combinations in which to remove constraints. We provide practical examples for problems concerning $P = 5$:
\begin{enumerate}
    \item $K = 4$, then $\binom{20}{14 + 4} = \binom{20}{18} = 2$, meaning that we have to remove 
\end{enumerate}

\section{Why this method works (WIP)}

\subsection{A general approach to relaxed formulations of overconstrained parametric polynomial systems}

Many estimation problems in vision are special cases of the following general problem. 
Suppose we are given an algebraic variety $X$, whose points consist of problem-solution pairs $(p,x)$ satisfying some set of $N$ polynomial constraints,
\begin{equation}\label{eq:def-X}
X = \{ (p, x) \in \CC^m \times \CC^n \mid f_1 (x;p) = \ldots = f_N (x; p)=0 \}. 
\end{equation}
We further assume that that variety $X$ is \emph{irreducible} (cf.~\cite[\S 5]{clo}.)
As we will explain in~\Cref{subsec:applying-framework}, this is often a reasonable assumption in practice.
Our task is to compute solutions $x$ for particular problem instances $p\in \CC^m.$ 
The set of solutions in question may be identified with the fiber $\pi^{-1} (p)$ over $p$ with respect to the projection map into the space of problems,
\begin{align}
\pi : X &\to \CC^m, \nonumber \\
\pi (p,x) &\mapsto p. \label{eq:pi-map}
\end{align}

We denote by $B_\pi$ the set of exactly-solvable problem instances: that is, 
\[
B_\pi = \{ p \in \CC^m  \mid \exists x \in \CC^n \text{ such that } \pi (p, x) = p \} .
\]
For many purposes, it is harmless to replace $B_\pi$ with its closure $\overline{B_\pi} \subset \CC^m$ in either the Zariski or Euclidean topologies.
This is an irreducible algebraic variety which contains $B_\pi $ as a dense subset.
Adapting the definition of~\Cref{plmp}, we say that $\pi$ represents a \emph{minimal problem} if the following three conditions hold:
\begin{enumerate}
    \item[(1)] Degrees of freedom are \emph{balanced} --- that is, $\dim X = \dim B_\pi.$
    \item[(2)] For a generic problem instance $p\in B_\pi,$ the fiber $\pi^{-1} (p)$ is zero-dimensional.
    \item[(3)] A generic problem instance is solvable---that is, $\overline{B_\pi } = \CC^m.$
\end{enumerate}

When dealing with \emph{overconstrained} problems, typically condition (1) above still holds, but not (2)--(3).
Nevertheless, we may hope to harness techniques for solving minimal problems by devising \emph{minimal relaxations} of overconstrained problems.
Here we explain one possible framework which can be used to analyze a broad class of minimal relaxations.

Let us restrict attention to a subset of equations vanishing on $X$, whose size is equal to the number of unknowns.
Without loss of generality, we may assume that these equations $f_1, \ldots , f_n$ appear first in our description of $X$ given in~\eqref{eq:def-X}. 
Now, write each $f_i$ as a polynomial in $x$ whose coefficients are polynomials in $p,$
\[
f_i (x;p) = \displaystyle\sum_{\alpha \in A_i} b_{i, \alpha } (p) x^\alpha ,
\]
where $A_i \subset \mathbb{Z}_{\ge 0}^n$ indexes the set of $x$-monomials appearing in $f_i.$
We may then the define the \emph{coefficient variety} associated to $X$ and the square system $f_1, \ldots , f_n$ to be the closed image of the map
\begin{align}
b_{\pi, f_1, \ldots , f_n} : \overline{B_\pi} &\to \CC^{\# A_1} \times \cdots \times \CC^{\# A_n} \nonumber\\
p &\mapsto \left( b_{i, \alpha} (p) \mid 1\le i \le n, \alpha \in A_i \right). \label{eq:b-map}
\end{align}
Let $B_{\pi , f_1, \ldots , f_n} \subset \CC^{\# A_1} \times \cdots \times \CC^{\# A_n}$ denote this coefficient variety. 
Notice that $b_{\pi, f_1, \ldots , f_n}$ is the restriction of a polynomial map on $\mathbb{C}^m.$ 
We are interested in cases where the closed image of this map and $B_\pi$ coincide.
This ``reparametrization" property, defined below, will allow us to construct \emph{parameter homotopies} that deform one problem instance into another.
\begin{enumerate}
    \item[(1')] \tim{Remove this --- replace the submersive condition} (Reparametrization) The image of the map $\CC^{m} \to \CC^{\# A_1} \times \cdots \times \CC^{\# A_n}$ is dense in $B_\pi .$
\end{enumerate}
Consider now $b_{f_1, \ldots , f_n} \circ \pi ,$ the composition of the maps~\eqref{eq:pi-map} and~\eqref{eq:b-map}.
Similar to our notation for $B_{\pi, f_1, \ldots  f_n},$ let $X_{\pi, f_1, \ldots , f_n}$ denote the image of $X$ under the projection 
\begin{align*}
    \operatorname{id}_{\CC^n} \times b_{f_1, \ldots , f_n} : \CC^n \times \CC^m \to \CC^n \times \CC^{\# A_1 } \times \cdots \times \CC^{\# A_n}.
\end{align*}
As a preliminary definition of minimal relaxation, we will require that the projection map
\[
\pi_{f_1, \ldots , f_n } : X_{\pi , f_1, \ldots , f_n} \to B_{\pi, f_1, \ldots , f_n}
\]
satisfies a generalization of the balancing condition (1) above.
\begin{enumerate}
\item[(2')] (Balancing conditions) $\dim X = \dim X_{\pi, f_1, \ldots, f_n} = \dim B_{\pi, f_1, \ldots , f_n}.$
\end{enumerate}
Finally, in order to make solving minimal relaxations practical, we also make the following assumption.
\begin{enumerate}
\item[(3')] (Sampling oracle) We can sample a generic problem-solution pair $(b_0, x_0) = (b_{f_1, \ldots , f_n} (p_0), x_0) \in X_{\pi, f_1, \ldots , f_n}$ with the property that the $n\times n$ Jacobian $(\displaystyle\frac{\partial f_j}{\partial x_j} (b_0, x_0))_{1\le i, j \le n }$ has full rank.
\end{enumerate}
\begin{definition}\label{def:minimal-relaxation}
Let $\pi : X \to \CC^m$ be a projection map as in~\eqref{eq:pi-map}, where $X$ is an algebraic variety as in~\eqref{eq:def-X} describing a set of problem-solution pairs.
Let $f_1, \ldots , f_n$ be a square system of equations vanishing on $X.$
We say that $f_1, \ldots , f_n$ determine a \emph{minimal relaxation} if assumptions (1')--(3') are satisfied for the projection map $\pi_{f_1, \ldots , f_n}$ defined above.
\end{definition}
Notice that the map $\pi_{f_1, \ldots , f_n}$ is surjective, hence (2') above is equivalent to requiring that the fibers of the map $\pi_{f_1, \ldots , f_n}$ are zero-dimensional, the natural analogue of (2).
As for property (3), usually $B_{\pi, f_1, \ldots , f_n}$ is \emph{not} dense in $\CC^{\# A_1} \times \cdots \times \CC^{\# A_n}.$

We stress that, although $f_1, \ldots , f_n$ are equations vanishing on $X_{\pi , f_1, \ldots , f_n}$, we have not assumed that they form a \emph{complete} set of equations.
From our assumptions so far, we may only conclude that $X_{\pi, f_1, \ldots , f_n}$ is an irreducible subvariety of the vanishing locus of $f_1, \ldots , f_n.$
In fact, neither a complete set of equations for $X$ nor for $X_{\pi, f_1, \ldots , f_n}$ is needed in practice to solve a minimal relaxation.
Finally, let us note that property (3') is enough to give the second equality in property (2').
To see this, note that the fiber $\pi_{f_1, \ldots , f_n}^{-1} (b_0)$ is a subset of the vanishing set of $f_1(x; p_0) = \ldots , f_n (x; p_0) = 0$. Since (1'') implies this vanishing set is zero-dimensional, the dimension of the fiber is at most $0$.
Thanks to our sampled point $(b_0, x_0) \in \pi_{f_1, \ldots , f_n}^{-1} (b_0),$ the fiber is non-empty, and hence also zero-dimensional. 

\textbf{Remark 1} \Cref{def:minimal-relaxation} is general enough to capture all of the minimal relaxations considered in this work.
However, we do not propose an all-encompassing theory of minimal relaxations.
For example, the ``Scranton" relaxation of reconstructing four points in three calibrated views used in~\cite{hruby-learning} is based on the introduction of a slack variable, and does not conform to our present framework in any obvious way.

\subsection{Applying the framework}

We now apply the framework of~\Cref{subsec:framework} to 3D reconstruction problems with partially-specified calibration data.
Suppose we wish to reconstruct $P$ points using $N$ cameras.
\tim{P or N?}
We assume that the space of camera $N$-tuples is parametrized by an algebraic variety $\mathcal{C}_N$ via a rational map
\begin{align*}
\mathcal{C}_N &\dashrightarrow (\mathbb{P} (\mathbb{C}^{3\times 4}))^N\\
c &\mapsto (P_1(c), \ldots , P_n (c)).
\end{align*}
For this parametrization of cameras, we define the joint camera map
\begin{align}
(\CC^3)^P  \times \mathcal{C}_N &\dashrightarrow (\CC^3)^{NP} \nonumber \\
(X_1, \ldots , X_P, c) &\mapsto \left( u_i(X, c) , v_i (X, c), \lambda_{ij} (X, c) \mid 1 \le i \le N, 1\le j \le P \right),
\label{eq:joint-camera-map}
\end{align}
where $u_{ij}$ and $v_{ij}$ record the pixel coordinates and $\lambda_{ij}$ the depths of a generic point in each image.
Explicitly,
\begin{align}
u_{ij} (X, c) &= \frac{ (P_i (c) \cdot X_j)[1,:]}{(P_i (c) \cdot X_j)[3,:]},\nonumber \\
v_{ij} (X, c) &= \frac{ (P_i (c) \cdot X_j)[2,:]}{(P_i (c) \cdot X_j)[3,:]},\nonumber \\
\lambda_{ij} (X, c) &= \displaystyle\frac{(P_i (c) \cdot X_j)[3,:]}{(P_1 (c) \cdot X_1)[3,:]}.
\label{eq:uvlambda-map}
\end{align}
Let $X'$ denote the graph of this map.
Projecting $X'$ away from the space of world points and subspace of camera intrinsic parameters in $\mathcal{C}_N,$ we obtain a variety $X$ of the form~\eqref{eq:def-X}, whose coordinates correspond to calibration matrices depths (solutions), and pixel coordinates (problem instances.)
The problems we consider are all minimal relaxations obtained from the space of problem-solution pairs $X.$
Having obtained solutions in depths, we can then recover rotations using formulas from~\cite{learning-hruby}. After this, translations can be recovered linearly---note that the normalization $\lambda_{11}=1$ in~\eqref{eq:uvlambda-map} fixes the global scale ambiguity.

{\small
\bibliographystyle{ieee_fullname}
\bibliography{refs, Andrea}
}
\end{document}


\maketitle

\section{Enumerating Autocalibration Minimal Relaxations}

\subsection{Line graph isomorphism}
The isomorphism class of a $4$-coloring is completely determined by an associated line graph $\LL(c)$ equipped with a suitable labeling. 
The vertices of $\LL(c)$ are simply the non-white edges, and an edge between two vertices of $\LL(c)$ exists whenever they share a vertex between $1$ and $N.$
We label each vertex $pq \in \LL(c)$ by its color $c(pq).$ 
Two isomorphic graphs have isomorphic line graphs; conversely, a classical theorem of Whitney~\cite{whitney1992congruent} implies that two connected graphs whose line graphs are isomorphic are themselves isomorphic, with the sole exception of the complete graph $K_3$ and the claw graph $K_{1,3}$ (see eg.~\cite[Theorem 8.3]{harary}.)
Although  $\LL(K_3) \cong \LL(K_{1,3})$, the original graphs $K_3$ and $K_{1,3}$ have a different numbers of edges. 
From this, it easily follows that we can decide whether or not two $4$-colorings $c_1,c_2$ are isomorphic: form the graphs $\LL(c_1)$ and $\LL(c_2)$ and decide if there exists an isomorphism that respects the labeling, and repeat this same procedure for $\LL(c_1), \LL(c_2 \circ \tau).$

\section{Experiments}

\begin{equation} 
\label{eq:experiments-delta-fg}
\Delta \mathtt{fg} = \frac{1}{2} \left( \frac{\left| \hat{\mathtt{f}} - \mathtt{f}_\mathrm{gt} \right|}{\mathtt{f}_\mathrm{gt}}  + \frac{\left| \hat{\mathtt{g}} - \mathtt{g}_\mathrm{gt} \right|}{\mathtt{g}_\mathrm{gt}} \right) \; ,
\end{equation}

\begin{equation} 
\label{eq:experiments-delta-uv}
\Delta \mathtt{uv} = \frac{1}{2} \left( \frac{\left| \hat{\mathtt{u}} - \mathtt{u}_\mathrm{gt} \right|}{\mathtt{u}_\mathrm{gt}}  + \frac{\left| \hat{\mathtt{v}} - \mathtt{v}_\mathrm{gt} \right|}{\mathtt{v}_\mathrm{gt}} \right) \; .
\end{equation}

{\small
\bibliographystyle{ieee_fullname}
\bibliography{refs, Andrea}
}